\documentclass[12pt,a4paper,twoside]{article}
\usepackage[utf8]{inputenc}
\usepackage[top=3cm,bottom=3cm,left=2.5cm,right=2.5cm]{geometry}
\usepackage[sf,bf]{titlesec} 

\title{{\LARGE\bf\sffamily How to Allocate Your Tokens? \\ \Large Scaling Laws with Training Steps and Batch Size \par}}
\newcommand{\footremember}[2]{%
    \footnote{#2}
    \newcounter{#1}
    \setcounter{#1}{\value{footnote}}%
}

\author{Fabian Schaipp\footremember{myemail}{Corresponding email: \url{fabian.schaipp@inria.fr}}
\footremember{inria}{Inria, \'{E}cole Normale Sup\'{e}rieure, PSL Research University, Paris}
}

\date{}

\usepackage{xcolor}
\usepackage{amsmath,amssymb,amsfonts,amsthm}
\usepackage{mathtools}
\usepackage{bm}
\usepackage{dsfont}
\usepackage[]{algorithmic, algorithm}
\usepackage{enumitem}
\usepackage{natbib}
\usepackage{verbatim}

\usepackage{multirow}
\usepackage{longtable}
\usepackage{booktabs}

\usepackage[hidelinks,pagebackref]{hyperref}
\renewcommand*{\backref}[1]{}
\renewcommand*{\backrefalt}[4]{{\footnotesize [%
		\ifcase #1 Not cited.%
		\or Cited on page~#2%
		\else Cited on pages #2%
		\fi%
		]}}
	
\usepackage[capitalize,nameinlink]{cleveref}
\usepackage{subcaption} 
\usepackage{caption}

\usepackage{url}
\usepackage{cite}

\newtheorem{theorem}{Theorem}

\newtheorem{definition}[theorem]{Definition}

\usepackage[parfill]{parskip}

\makeatletter
\def\thm@space@setup{%
	\thm@preskip=\parskip \thm@postskip=0pt
}
\makeatother

\let\temp\phi
\let\phi\varphi
\let\varphi\temp

\let\temp\varepsilon
\let\epsilon\varepsilon
\let\varepsilon\temp

\newcommand{\N}{\mathbb{N}}

\newcommand{\E}{\mathbb{E}}

\newcommand{\Adam}{{\texttt{Adam}}}
\newcommand{\AdamW}{{\texttt{AdamW}}}

\newcommand{\li}{\textbf{Li}}
\newcommand{\oellm}{\textbf{OpenEuroLLM}}

\newcommand{\ttl}{{\texttt{2TL}}}
\newcommand{\tttl}{{\texttt{3TL}}}

\newcommand{\ie}{i.e.\ }

\definecolor{kleinblue}{RGB}{0, 47, 167}
\definecolor{royalblue}{RGB}{0, 33, 115}
\hypersetup{
	colorlinks = True,
	linkcolor = black,
	citecolor=royalblue,
	urlcolor=royalblue
} 

\definecolor{todored}{RGB}{189, 30, 30}

\definecolor{nypink}{RGB}{216, 131, 115}
\definecolor{commblue}{RGB}{0, 66, 102}

\definecolor{takeawaycolor}{RGB}{155, 157, 137}
\definecolor{caveatcolor}{RGB}{224, 122, 95}
\definecolor{StrongRed}{RGB}{230, 57, 70}
\definecolor{OtherBlue}{RGB}{29, 112, 175}

\usepackage[colorinlistoftodos]{todonotes}

\newcommand{\takeaway}[1]{\todo[inline,bordercolor=takeawaycolor,backgroundcolor=takeawaycolor!20,linecolor=takeawaycolor]{\textbf{Takeaway: }#1}}
\newcommand{\caveat}[1]{\todo[inline,bordercolor=caveatcolor,backgroundcolor=caveatcolor!20,linecolor=caveatcolor]{\textbf{Caveat: }#1}}

\begin{document}
\maketitle

\begin{abstract}
    We propose a scaling law that takes into account model size and training data while explicitly splitting the latter into training steps and batch size (called \emph{three-term law}). Fitting the proposed law on a large set of training runs, we find that it correctly recovers the scaling of the optimal batch size. Moreover, because it makes use of training runs with suboptimal batch size, our proposed law can be robustly fit with a significantly smaller amount of training runs.
    We further show that the three-term law can be used to derive scaling laws for suboptimal batch sizes, and that it matches previous empirical findings related to the critical batch size.
\end{abstract}

\section{Introduction}

The field of deep learning and specifically large language models (LLMs) has seen an enormous progress over the last few years. Much of this progress has been attributed to ``simply" scaling up, both in terms of model size and data used for training. Improvements in model performance often follow predictable trends, called \emph{scaling laws}, which have been found in the context of LLM pre-training \citep{Kaplan2020,Hoffmann2022}, but also in many other areas of deep learning such as vision \citep{Zhai2022}, weather forecasting \citep{Bodnar2024}, or protein modeling \citep{Lin2023}.

\paragraph{Scaling laws for model size and data.} In the context of training large language models, scaling laws classically refer to functional forms that allow to infer the optimal allocation of compute into model size $N$ and training examples $D$. Seminal works by \citet{Kaplan2020} and \citet{Hoffmann2022} show that the test loss predictably decreases when increasing $N$ or $D$. A widely-used technique to obtain a scaling law, known as Chinchilla Approach 3, is to model the (final test) loss as a sum of power-laws in $N$ and $D$, \ie
\begin{align}\label{eqn:scaling-law}
    \mathcal{L}(N,D) = E + \frac{A}{N^\alpha} + \frac{B}{D^\beta}.
\end{align}
The parameters $(E,A,B,\alpha,\beta)$ are fitted from a set of training runs. One can then derive the optimal model size from \eqref{eqn:scaling-law} for a given compute constraint, and potentially extrapolate this to values of $(N,D)$ outside of the experimentally tested ranges. On the other hand, several works show that the fitting procedure itself is delicate, and design choices in both the training procedure as well as the fitting techniques can impact the result \citep{Besiroglu2024,Li2025a}. Further, it is not guaranteed that the law generalizes well, which however is crucial to derive optimal configurations for large-scale training runs.

\paragraph{Hyperparameter scaling laws.} While scaling laws of the Chinchilla form can inform compute-optimal allocation of $N$ and $D$, they do not explicitly account for training hyperparameters. To address this, several recent works derive scaling laws for the optimal learning rate and/or batch size as a function of $(N,D)$ \citep{DeepSeekAI2024,Li2025,Ruette2026}. Typically, these works simply assume a power-law relationship for the hyperparameter of interest (e.g.\ the optimal batch size as a function of $D$) and fit the coefficients to data from training runs. However, these approaches are not directly compatible with laws of form \eqref{eqn:scaling-law} as they do not model the loss value. 

A different line of work is based on the concept of the \emph{critical batch size} \citep{McCandlish2018,Shallue2018}: it describes the phenomenon that the number of steps $K$ required to reach a target loss, as a function of the batch size, will at some point decrease much slower than inversely linear. \citet{McCandlish2018} model this with the equation
\begin{align}\label{eqn:critical-batch-size}
    (K/K_{\min} - 1) (D/D_{\min} - 1) = 1.
\end{align}
Here, $K_{\min}$ (resp.\ $D_{\min}$) is the minimum number of steps (resp.\ number of tokens) required to reach the target loss, and can be fit empirically. The critical batch size is then defined as $D_{\min}/K_{\min}$. \citet{Bergsma2025} establish a connection to weight decay. In an earlier work, \citet{Kaplan2020} use the same model of critical batch size to relate the loss to the number of training steps. 

A central issue with \eqref{eqn:critical-batch-size} is that it implies an \emph{optimal batch size of one} \citep{Bergsma2025}; however, this is in conflict with the empirical situation when using \AdamW{}, where the optimal batch size has been found to scale with the available token budget $D$ \citep{Porian2024,Li2025}.

\paragraph{Scaling laws and optimization theory.} From a theoretical point of view, the relationship between loss, batch size and number of training steps is inherently related to optimization theory. Recent works by \citet{Shulgin2026,Islamov2026} derive hyperparameter scaling laws directly from convergence bounds for stochastic conditional gradient methods. Moreover, it has been shown previously that hyperparameter tuning/transfer for LLM training can be informed by optimization theory, for example in the context of learning-rate schedules \citep{Schaipp2025} or of weight decay \citep{Wang2025}.

\section{Overview}\label{sec:overview}

\paragraph{Our proposed laws.} Here, we propose to model the loss as a power-law function of $(N,M,K)$ where $M$ is the batch size in tokens and $K$ is the number of training steps, 
that is,
\begin{align*}
    \mathcal{L}(N,M,K) = E + \frac{A}{N^\alpha} + \frac{B}{M^\beta} + \frac{C}{K^\gamma}.
\end{align*}
This has several natural advantages:
\begin{enumerate}[label=(\roman*)]
    \item Our law brings together the Chinchilla form \eqref{eqn:scaling-law} with scaling laws for the optimal batch size. Under a constrained data budget $D=MK$, our law implies a scaling rule for the optimal batch size with $D$, while at the same time collapsing to a Chinchilla form when using the optimal batch size.
    \item The proposed form is also closely connected to and inspired by theoretical results from stochastic optimization, see \citet{Shulgin2026,Kovalev2025}. This allows to bridge from empirical scaling analysis to a more theoretical understanding.
    \item The proposed law can be fit with runs from suboptimal batch sizes, which (as we will show) drastically reduces the number of training runs needed for fitting.
    \item While previous scaling laws only model the performance with \emph{optimal} hyperparameters, our formulation describes performance also in the suboptimal batch size regime. This can be important in practice when facing hardware constraints.
\end{enumerate}

\paragraph{Summary of our findings.} 

We fit our proposed laws on two datasets of training runs of (dense) LLMs, from here on referred to as \li{} \citep{Li2025} and \oellm{} \citep{OpenEuroLLM2026}. Both datasets cover multiple model sizes, token budgets and batch sizes.\footnote{The \oellm{} dataset is not yet public at the time of writing. We will make our codebase for reproducing all experiments public once the \oellm{} dataset has been released.}

\begin{enumerate}[label=(\roman*)]
    \item Our law results in an implied optimal batch size scaling that is consistent with previous hyperparameter scaling laws that do not model the loss value (see \cref{sec:experiments:compare-2tl-3tl}). In particular, we find that with our formulation two batch sizes per $(N,D)$ suffice to robustly find this law (instead of doing a full sweep); this reduces the number of training runs needed to 28\% (see \cref{sec:saving-compute}).
    \item By construction, our proposed law results in non-trivial optimal batch sizes (in contrast to previous formulations, as mentioned above) and that are independent of model size, matching the empirical results of  \citet{Li2025}. At the same time, it describes how the critical batch size scales with $N$ and/or $D$ consistently with findings by \citet{Zhang2025} (see \cref{sec:cbs-analysis}).
    \item For situations where the optimal batch size might be infeasible due to practical constraints, we derive scaling laws for $\epsilon$-suboptimal batch sizes that generalize well to out-of-sample token budgets (see \cref{sec:suboptimal-regime}).
\end{enumerate}

\paragraph{Notation.}

The table below summarizes the most important notation used throughout the document. Note that it holds $D=MK$.

\begin{table}[H]
    \label{tab:notation}
    \centering
	\begin{tabular}{ | c | c | } 
		\hline
		\multicolumn{2}{|c|}{Main quantities} \\
		\hline \hline
        $\mathcal{L}$ & Loss value \\
		$N$ & Number of parameters (model size) \\
		$D$ & Size of training data (in tokens) \\
		$b$ & Batch size \\
        $s$ & Sequence length \\
		$K$ & Number of training steps \\
		$M$ & Batch size (in tokens). It holds $M=bs$.\\
        $\mathcal{C}$ & Compute (in flops). Usually modeled as $\mathcal{C}=6ND$. \\
		\hline
	\end{tabular}
\end{table}

\section{Scaling Laws with Training Steps and Batch Size}\label{sec:bs-step-scaling-laws}

Recall the Chinchilla law proposed by \citet{Hoffmann2022}: for $A,B,E >0$ and $\alpha, \beta > 0$ let the loss be parametrized as
\begin{align*}
    \mathcal{L}(N,D) = E + \frac{A}{N^\alpha} + \frac{B}{D^\beta}.
\end{align*}
Here, $\mathcal{L}$ usually refers to the test loss at the end of training (or a smoothed version of it). Using the training runs from \citet{Hoffmann2022}, but with a more precise fitting procedure, \citet{Besiroglu2024} report the law
\begin{align}\label{epochai-law} \tag{\text{EpochAI}}
    \mathcal{L}(N,D) = 1.8172 + \frac{482.01}{N^{0.3478}} + \frac{2085.43}{D^{0.3658}}.
\end{align}

The above scaling law does \textbf{not} take into account the batch size, which however plays a crucial role for training performance \citep{Bergsma2025,Zhang2025}. In particular, \citet{Hoffmann2022} report only that they used ``well-tested heuristics'' for the choice of batch size, but they do not study its effect on the scaling analysis. 

Here, we propose to take into account how the token budget $D$ \textbf{is allocated into} training steps $K$ and batch size $b$. We first describe two similar approaches to do so.

\paragraph{Approach I: A three-term law.}

Following the power-law approach, we propose the functional form 
\begin{align}\label{eqn:three-term-law}
    \mathcal{L}(N,M,K) = E + \frac{A}{N^\alpha} + \frac{B}{M^\beta} + \frac{C}{K^\gamma},
\end{align}
where $(E,A,B,C,\alpha,\beta,\gamma)$ are fittable parameters. This has the advantage that for optimal batch size choice it reduces automatically to the original Chinchilla form \eqref{eqn:scaling-law}: to see this, minimize \eqref{eqn:three-term-law} with respect to $M$, subject to $D=MK$. The optimal batch size is
\begin{align}\label{eqn:opt-batch-size}
    M^\star = \Big[\frac{\beta B}{\gamma C}\Big]^{\frac{1}{\beta + \gamma}} D^\frac{\gamma}{\beta + \gamma}=:GD^\frac{\gamma}{\beta + \gamma}.
\end{align}
In particular, the optimal batch size is independent of the model size $N$. Plugging back $M^\star$ and $K^\star = D/M^\star$ into \eqref{eqn:three-term-law} gives 
\begin{align}\label{eqn:three-term-optimal}
    \mathcal{L}(N,D) = E + \frac{A}{N^\alpha} + \frac{\hat B}{D^\tau}, \quad \tau:= \frac{\beta \gamma}{\beta + \gamma}, \quad \hat{B} := BG^{-\beta}+ CG^{\gamma}.
\end{align}
Hence, under \eqref{eqn:three-term-law}, the optimal way to split a fixed token budget $D$ into $(K, b)$ recovers the form \eqref{eqn:scaling-law} by re-parameterizing $(B, \beta) \to (\hat B, \tau)$.

Equation \eqref{eqn:three-term-law} is structurally very similar to convergence bounds from stochastic optimization. Let $(x_k)_{k \in \N}$ be the iterates of the stochastic conditional gradient method wrt.\ a general norm. Assuming the $\mu$-KL condition for the loss function (see \citet{Islamov2026}), for fixed batch size $M$ (or $b$) and training steps $K$, under the optimal learning rate $\eta^\star$, \citet[Thm.\ 1]{Shulgin2026} (derived from \citet{Kovalev2025}) states that
\begin{align*}
    \min_{1\leq k \leq K} \E[\mathcal{L}(x_k)]  \lesssim \mathcal{L}_\star +  \frac{1}{\sqrt{M}} + \frac{1}{\sqrt{K}},
\end{align*}
where $\lesssim$ denotes that the bound holds up to some multiplicative constant for each term on the right-hand side. Comparing to \eqref{eqn:three-term-law}, the optimal loss is parametrized by $E+\frac{A}{N^\alpha}$, and for the other terms the powers are relaxed from $1/2$ to $(\beta,\gamma)$.

\paragraph{Approach II: Model-specific two-term laws.}
The functional form \eqref{eqn:three-term-law} implicitly assumes that the model size does not impact the coefficients $(B,C)$ and $(\beta, \gamma)$, and therefore by construction $M^\star$ is independent of $N$.
While this implicit assumption is supported by previous experimental results on LLMs \citep{Li2025,Zhang2025}, as an alternative, we can \textbf{fix the model size} and fit the functional form 
\begin{align}\label{eqn:two-term-law}
    \mathcal{L}(M,K) = E + \frac{B}{M^\beta} + \frac{C}{K^\gamma},
\end{align}
where again $(E,B,C,\beta,\gamma)$ are fittable parameters. We can fit this form (independently) for multiple $N$,  a priori allowing for different values of $(B, C, \beta,\gamma)$ across model sizes.

\paragraph{Terminology.} From now on, we refer to laws of the form \eqref{eqn:two-term-law} as \emph{two-term law}, and to laws of the form \eqref{eqn:three-term-law} as \emph{three-term law}. When convenient, we will use the abbreviations \ttl{} and \tttl{} respectively.

\section{Experiments}

We fit scaling laws of the form \ttl{} and \tttl{} on two datasets which contain training logs for multiple combinations of $(N,M,K)$  using \texttt{AdamW} \citep{Loshchilov2019}, and which include a learning-rate sweep for each combination. We refer to the two datasets as \li{} and \oellm{}; details are described in \cref{sec:app:experiment-setup} in the Appendix, in particular see \cref{tab:dataset-overview}. Here, we focus on the \li{} dataset; the results for \oellm{} are deferred to the Appendix. For each $(N,M,K)$, we choose the smallest final smoothened test loss value across learning rates. We form a validation set by collecting the datapoints from the largest token budget $D$ for each individual $N$; the remaining datapoints are used as training set for fitting the law. We usually mark points from the validation set with a dashed border in our plots. 
See \cref{sec:app:experiment-setup} for additional details and \cref{fig:data-overview-li,fig:heatmaps-li} for an overview of the \li{} dataset. 

\paragraph{Fitting procedure.} If not explicitly mentioned otherwise, we use a standard fitting procedure minimizing the Huber loss with \texttt{L-BFGS-B} from multiple initializations, and five-fold cross-validation for each law. For the Huber loss we use $\delta=10^{-3}$; for a detailed description and additional ablations on these choices, we refer to \cref{sec:app:experiment-setup} and references therein. All fitted parameter values can be found in \cref{sec:app:coefficients} in the Appendix. \textbf{Note that} whenever we report a single number for a parameter, it is the average over the five cross-validation runs.

\subsection{Comparison of Approach I and II}\label{sec:experiments:compare-2tl-3tl}

A priori, it is not clear which of the two approaches described in \cref{sec:bs-step-scaling-laws} is preferable.
Before doing any experimental comparison, we want to mention the main differences of \ttl{} and \tttl{} that arise from their definition. First, for \tttl{} we will usually have more datapoints per (fittable) parameter, as we can use training runs across all model sizes $N$. For example, for the \li{} dataset, we have 246 samples to fit seven parameters for \tttl{}, while for \ttl{} we have 13-60 samples to fit five parameters. Consequently, we expect that the in-sample fitting error of \ttl{} will be smaller. Second, based on the results of \citet{Li2025a}, we would expect that \tttl{} is more delicate/unstable to fit, given that it has two more fittable parameters.

We compare both approaches with respect to the following evaluations: (i) quality of fit, (ii) consistency of the scaling coefficients $(\beta,\gamma)$ and (iii) consistency of the resulting optimal batch size $M^\star$ from \eqref{eqn:opt-batch-size} with previously reported results.
We fit a \ttl{} for each model size $N$ (using only the respective subset of datapoints), and a single \tttl{} 
using the union of all datapoints (across $N$).

\begin{figure}[t]
    \centering
    \begin{subfigure}[b]{0.49\columnwidth}
        \includegraphics[width=0.99\linewidth]{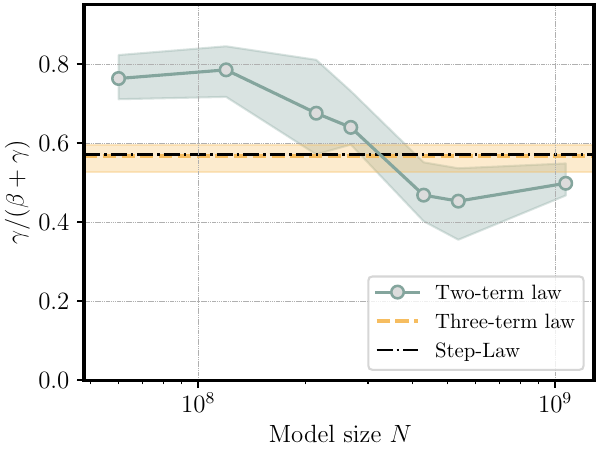}
    \end{subfigure}
    \begin{subfigure}[b]{0.49\columnwidth}
        \includegraphics[width=0.99\linewidth]{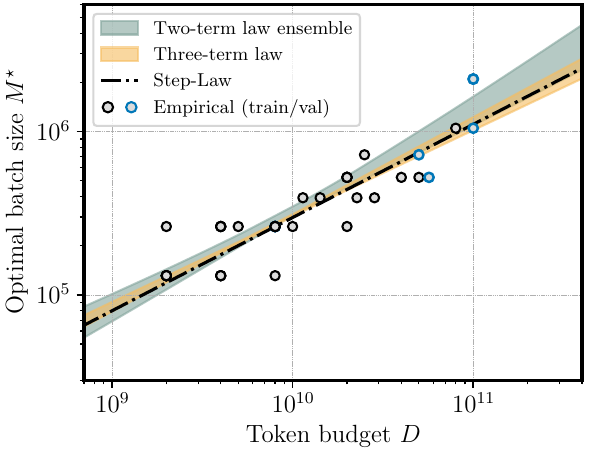}
    \end{subfigure}
    \caption{\textbf{(Left)} Estimates for $M^\star$-scaling coefficient $\frac{\gamma}{\beta + \gamma}$ for \tttl{} and each \ttl{}. Shaded area depicts min and max over five cross-validation fits. \textbf{(Right)} Implied scaling of $M^\star$ according to \eqref{eqn:opt-batch-size}. Shaded area depicts min and max over cross-validation. Dots show the empirically best batch size from the train (\textit{black}) and validation split (\textit{blue}).}
    \label{fig:scaling-law-comparison-li}
\end{figure}

\begin{figure}[t]
    \centering
    \begin{subfigure}[b]{0.49\columnwidth}
        \includegraphics[width=0.99\linewidth]{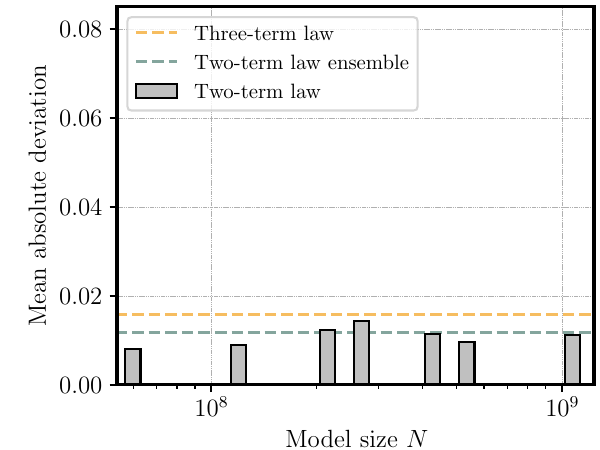}
    \end{subfigure}
    \begin{subfigure}[b]{0.49\columnwidth}
        \includegraphics[width=0.99\linewidth]{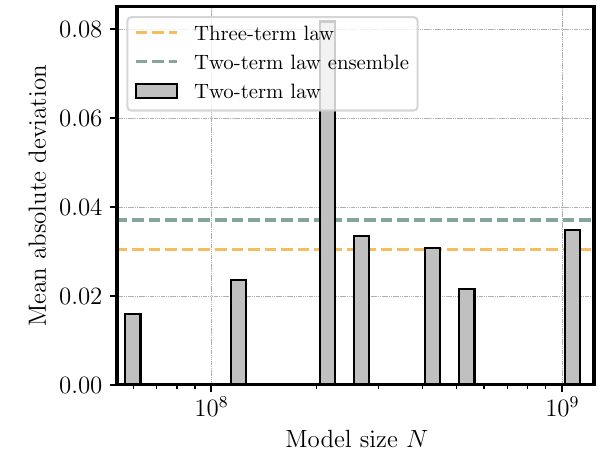}
    \end{subfigure}
    \caption{MAD comparison of \ttl{} and \tttl{} on train \textbf{(left)} and validation \textbf{(right)} split.}
    \label{fig:mad-comparison-li}
\end{figure}

For (i), we compute the mean absolute deviation (MAD) for the \tttl{} as well as for the \ttl{}, where each \ttl{} is evaluated on the subset of datapoints respective to the model size. We then compute the average MAD across all \ttl{}, weighted by the sample size, which we call the \emph{two-term law ensemble}. This is done separately for training and validation set (see \cref{fig:mad-comparison-li}). For (ii) and (iii), we plot the estimated values of $(\beta, \gamma)$ as well as the resulting formula for $M^\star$ from \eqref{eqn:opt-batch-size}. We can directly compare the latter to the scaling law for optimal batch size reported by \citet{Li2025}, which was obtained from the same dataset and which we refer to as \textbf{Step-Law} (see \cref{fig:scaling-law-comparison-li}).

\textit{Discussion:} As expected, \cref{fig:mad-comparison-li} shows that the in-sample fit of \ttl{} is better; however, the out-of-sample error of \tttl{} is slightly smaller. This indicates that despite having two additional parameters in \tttl{}, our fitting procedure is sufficiently robust thanks to multiple initializations and bootstrap aggregation. Moreover, \cref{fig:scaling-law-comparison-li} shows that also in terms of consistency of the implied scaling of $M^\star$ the three-term law approach is preferable. In particular, its resulting scaling of $M^\star$ almost perfectly overlaps with the one by  \citet{Li2025}. 

\takeaway{The three-term law approach achieves an overall slightly better out-of-sample fit than the two-term laws. Its implied scaling for the optimal batch size $M^\star \sim D$ is consistent with previous analyses (see \cref{tab:literature-bs-scaling}).
}

\paragraph{Consistency across datasets.} We run the same analysis on the \oellm{} dataset, see \cref{sec:app:oellm-results} in the Appendix. On \oellm{}, both approaches again lead to a good fit. The scaling of the optimal batch size is relatively similar to the \li{} dataset, albeit slightly steeper. However, we also observe the caveat described below.
\caveat{The fitted parameter values for $(E,A,\alpha)$ are quite different between \li{} and \oellm{}; in particular, the fit for \li{} is $E\approx 0$, indicating no irreducible loss which is in conflict to the non-zero entropy of language. This suggests that the effect of the model size is not perfectly reflected in \tttl{}. 
}
We hypothesize that the reason for scaling inconsistencies between the two datasets is due to different training setups. For example, \oellm{} uses different $\beta_2$ values in \Adam{} for the larger models, which is known to have an impact on scaling laws \citep{Porian2024}. This might also explain that the \ttl{} has a better out-of-sample error than \tttl{} for \oellm{}, as \ttl{} has more flexibility with respect to the impact of $N$. We also remark that imposing small regularization on $\log(E)$ can alleviate the above caveat (see \cref{sec:app:additional-observations}).
\begin{table}[h]
    \caption{Comparison of \tttl{} to batch size scaling laws from the literature.}
    \label{tab:literature-bs-scaling}
    \centering
    \begin{tabular}{|c|c|c|}
    \hline
         Reference & Scaling & Comment \\ \hline \hline
         \citep{Li2025} & $M^\star = 0.58\cdot D^{0.571}$ & Referred to as Step-Law \\
         \citep{Bergsma2025} & $M^\star = (0.0306 \cdot s) \cdot D^{0.383}$  & \\
         \citep{DeepSeekAI2024}& $M^\star = 0.086 \cdot D^{0.688}$ & From $M^\star \propto C^{0.327}$, $D^\star \propto C^{0.475}$ \\[3pt]
         \hline \hline
         (\tttl{}, \li{} dataset) \textbf{(ours)} & $M^\star = 0.667 \cdot D^{0.566}$ & From \eqref{eqn:opt-batch-size} \\[1pt]
         \hline
    \end{tabular}
\end{table}
\subsection{Compute Savings Using the Three-term Law}\label{sec:saving-compute}

Fitting a scaling law for $M^\star$ with the approach of \citet{Li2025} imposes massive computational costs, as it requires to obtain the optimal batch size for a set of different token budgets $D$ (and possibly also varying the model size $N$). \citet{Li2025} report that producing their entire set of training runs consumed nearly one million NVIDIA H800 GPU hours. Recall that \citet{Li2025} fits a law of form
\begin{align}\label{eqn:direct-fit-Mstar}
    M^\star = \frac{\tilde{A}}{D^{\tilde{\alpha}}}
\end{align}
directly on observations of $(M^\star, D)$. For a single observation $(M^\star, D)$, a full batch size sweep is needed to determine $M^\star$ (in the \li{} dataset, concretely we have 5-10 batch sizes per sweep). In contrast, our three-term law makes explicit use of observations from \emph{suboptimal} batch sizes; we will show that this allows to obtain the same scaling law of $M^\star$ while saving a substantial amount of training runs/compute.

\textit{Setup:} We mask the original dataset (containing the full batch size sweep), such that for each combination of $(N,D)$ \textbf{only \{one/two/three\} batch sizes are randomly selected} (see \cref{fig:data-overview-li-mask3} for an illustration). For the \li{} dataset, this shrinks the number of training runs required/available for the fit to \{14/28/42\} per-cent (see \cref{tab:mstar-laws-with-mask}). We then fit the three-term law on this reduced dataset.
As comparison, we fit \eqref{eqn:direct-fit-Mstar} with $M^\star$ being the batch size (after applying the mask as described above) with the best loss, for each $(N,D)$ separately.\footnote{
    This is similar to the procedure described by \citet{Li2025}. Alternatively, one could first fit a quadratic (or similar function) to the batch size sweep, and then read off $M^\star$ as the minimum. However, this is infeasible/unstable with $\leq 3$ points available. We try deriving $M^\star$ from a quadratic fit when having 4 points per sweep, but this does not improve the result, rather the opposite (the resulting scaling is $0.02\cdot D^{0.738}$).
} We also use five-fold cross-validation to fit \eqref{eqn:direct-fit-Mstar}.

\begin{figure}[t]
    \centering
    \begin{subfigure}[b]{0.49\columnwidth}
        \includegraphics[width=0.99\linewidth]{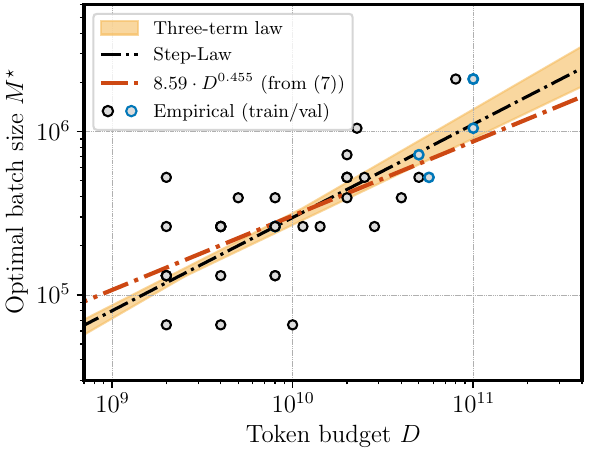}
        \caption{Dataset reduced to 42\%}
    \end{subfigure}
    \begin{subfigure}[b]{0.49\columnwidth}
        \includegraphics[width=0.99\linewidth]{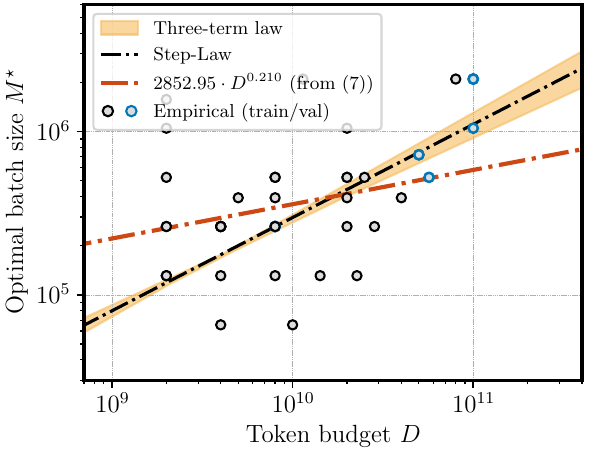}
        \caption{Dataset reduced to 28\%}
    \end{subfigure}
    \caption{Fitting on a reduced dataset, with only 3 values of $b$ per sweep \textbf{(left)} and 2 values \textbf{(right)}. Step-Law can be considered the oracle law here, as it was fit on the unreduced dataset (train plus val). In both cases, the implied scaling of $M^\star$ of the three-term law stays close to Step-Law, and generalizes better to large $D$ than the direct fit (in \emph{red}). The gray dots mark the empirically best batch size for each $(N,D)$ on the reduced dataset (for the train split). 
    } 
    \label{fig:masked-optimal-bs-scaling}
\end{figure}

\textit{Discussion:}
We make two main observations: (i) Already with the full batch size sweep, \eqref{eqn:direct-fit-Mstar} is unstable to removing the validation split. A direct fit on the full dataset (train and val) with our code gives $M^\star = 0.47 \cdot D^{0.584}$, essentially the same as Step-Law. Removing the validation split, we already get a quite different scaling of $M^\star = 6.29 \cdot D^{0.468}$. (ii) The three-term law results in almost identical scaling for $M^\star$, even when reducing the batch size sweep to two values, hence reducing the number of required training runs to 28\%. Directly fitting \eqref{eqn:direct-fit-Mstar} in contrast is highly unstable in this regime, and generalizes badly to higher (unseen) token budgets (see \cref{fig:masked-optimal-bs-scaling}). When masking to only one batch size per sweep, the results of both approaches are very distinct to the original law. 

\begin{table}[h]
    \centering
    \renewcommand{\arraystretch}{1.2}
    \begin{tabular}{|c||c|c||c|}
        \hline
         Sweep size for $b$ & \tttl{} & Direct fit \eqref{eqn:direct-fit-Mstar} & Samples/Training runs \\ \hline \hline
         Full & $M^\star = 0.67 \cdot D^{0.566}$ & $M^\star = 6.29 \cdot D^{0.468}$ & 246 \\
         3 values & $M^\star = 0.48 \cdot D^{0.580}$ & $M^\star = 8.59 \cdot D^{0.455}$ & 102 \\
         2 values & $M^\star = 0.84 \cdot D^{0.555}$ & $M^\star = 2852.95 \cdot D^{0.210}$ &  68 \\
         1 value & $M^\star = 5.92 \cdot D^{0.475}$ & $M^\star = 3.61 \cdot D^{0.514}$ &  34 \\
         \hline
         \multicolumn{4}{|c|}{\textbf{Step-Law}  \citep{Li2025}: $M^\star = 0.58\cdot D^{0.571}$} \\ \hline
    \end{tabular}
    \caption{Laws for $M^\star \sim D$ using \tttl{} vs.\ a direct fit, for different masked versions of the \li{} dataset. When having only two or three runs/batch sizes per $(N,D)$, \tttl{} still results in essentially the same law, whereas the direct fit deviates. Note that the direct fit only uses the train split, therefore the difference to the Step-Law.}
    \label{tab:mstar-laws-with-mask}
\end{table}

\subsection{Performance with Suboptimal Batch Sizes}\label{sec:suboptimal-regime}

In practice, understanding how the \emph{optimal} batch size scales with $N$ and $D$ might not be enough. In case of hardware constraints, it is mandatory to model the performance of models trained with \emph{suboptimal} batch size. 
The three-term law form has the evident appeal that it also predicts model performance for suboptimal allocation of tokens into steps and batch size.\footnote{This is similar to how the Chinchilla Approach 3 has the advantage that it describes suboptimal allocation of compute into model size or token budget.} In short, the goal of this section is to answer the following question:
\begin{quote}
    \textit{What is the interval of sub-optimal batch sizes $[b_{\min}, b_{\max}]$ such that at most 5\% of compute is wasted, and how does it scale with $D$?}
\end{quote}

\begin{figure}[t]
    \centering
    \begin{subfigure}[b]{0.49\columnwidth}
        \includegraphics[width=0.99\linewidth]{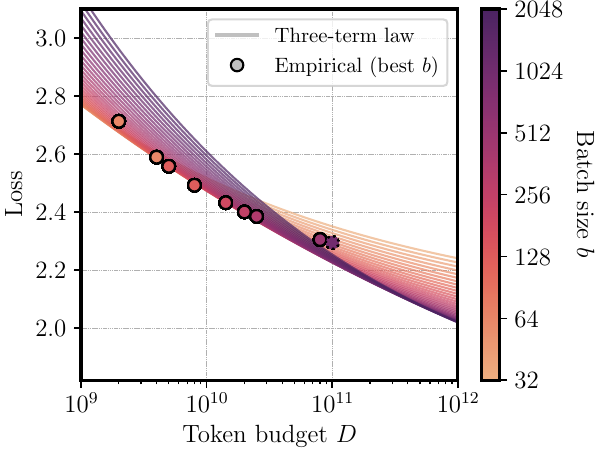}
    \end{subfigure}
    \begin{subfigure}[b]{0.49\columnwidth}
        \includegraphics[width=0.99\linewidth]{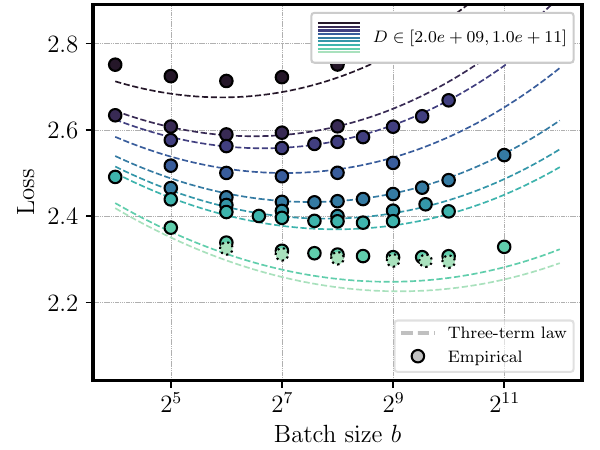}
    \end{subfigure}
    \caption{$N=268$M. \textbf{(Left)} While the three-term law \eqref{eqn:three-term-law} accurately predicts optimal batch size, its predicted \emph{loss value} for very large/small token budgets deviates from the empirical value. \textbf{(Right)} Empirical and predicted loss value across batch size $b$. Again, for very large/small token budgets the accuracy of the three-term law degrades. Dashed border marks datapoints not used for fitting \tttl{}.
    } 
    \label{fig:suboptimal-batch-size-3tl}
\end{figure}

\paragraph{Limitations of the three-term law.} We first evaluate the fitted \tttl{} on a range of batch sizes, and compare the predicted loss values to the true ones (see \cref{fig:suboptimal-batch-size-3tl}). While the optimal batch size is predicted well across all token budgets (\cref{fig:suboptimal-batch-size-3tl}, left), the three-term laws fails to accurately predict the loss value at the boundaries of the empirically covered range of $D$ (\cref{fig:suboptimal-batch-size-3tl}, right). 
This is not surprising: note that for the three-term law, we fit 246 data points with seven parameters; therefore, we can not expect a perfect fit.

\caveat{Due to its underparametrization, the three-term law can not fit loss values accurately enough to robustly infer performance at suboptimal batch sizes.}

\paragraph{Fitting in two stages.} To improve the fitting quality, we can fit a functional form $\mathcal{L} \sim b$ only on a subset of the data. We use the three-term law \eqref{eqn:three-term-law} as prior to choose such a functional form. Assume some fixed $N$ and $D$ and fixed sequence length $s$. Then, \eqref{eqn:three-term-law} simplifies to 
\begin{align*}
    \mathcal{L} = E + \frac{A}{N^\alpha} + (B s^{-\beta}) b^{-\beta} + (CD^{-\gamma}s^\gamma) b^{\gamma}.
\end{align*}

Based on the parameters of the fitted \tttl{}, we make the simplifying assumption $\gamma \approx \beta$ (we will also see that this is sufficiently expressive to give an almost perfect fit). Based on the above, we then fit the form
\begin{align}\label{eqn:suboptimal-bs-form}
    \mathcal{L}(b) = \tilde{E} + \tilde{A} b^{-\tilde{\alpha}} + \tilde{B} b^{\tilde{\alpha}}.
\end{align}

As a first stage, we fit $(\tilde{E}, \tilde{A}, \tilde{B}, \tilde{\alpha})$ from \eqref{eqn:suboptimal-bs-form} separately for each $(N,D)$.\footnote{
Here we perform a simple non-linear least squares fit using \texttt{scipy.optimize.curve\_fit}.}
This two-stage fitting procedure, where we reduced the number of parameters by assuming $\gamma \approx \beta$, has also been  recommended in the survey of \citet{Li2025a}.

Now, let us define the notion of $\epsilon$-suboptimal batch size:
\begin{definition}
    Let $\epsilon > 0$, and let $b^\star$ be the minimizer of \eqref{eqn:suboptimal-bs-form} (for a fixed $D$ and $N$). Since \eqref{eqn:suboptimal-bs-form} is unimodal in $b$, we can define $[b_{\min}, b_{\max}]$ to be the interval of $\epsilon$-suboptimal batch sizes such that $\mathcal{L}(b_{\min}) = \mathcal{L}(b_{\max}) = \mathcal{L}(b^\star) + \epsilon$.
\end{definition}
Here, we set $\epsilon$ to the loss difference from the law \eqref{epochai-law} evaluated at $(N,D)$ and $(N, 0.95\cdot D)$, that is, we allow a $5\%$ suboptimality in terms of compute.
From \eqref{eqn:suboptimal-bs-form}, we can then easily read off, for each $(N,D)$, the interval of $\epsilon$-suboptimal batch sizes.

As a second stage, we fit a power-law $b_{\min/\max} = \Upsilon /  D^{\nu}$, where $(\Upsilon, \nu)$ are fitted. Here, for each model size $N$ we keep the largest three token budgets as held-out validation set, and only use values of $D$ where the empirically optimal batch size does not lie on the boundary of the sweep.

\begin{figure}[t]
    \centering
    \begin{subfigure}[b]{0.49\columnwidth}
        \includegraphics[width=0.99\linewidth]{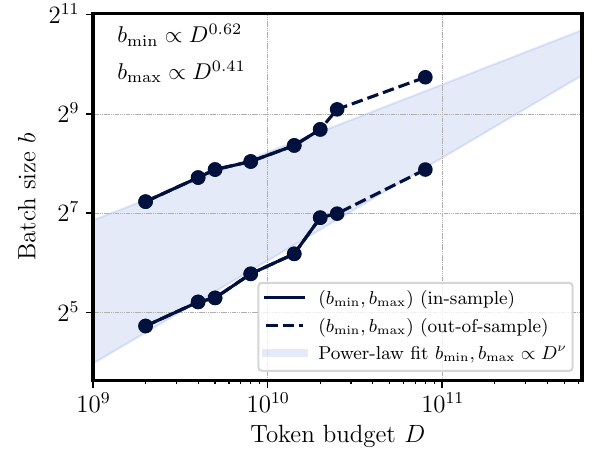}
    \end{subfigure}
    \begin{subfigure}[b]{0.49\columnwidth}
        \includegraphics[width=0.99\linewidth]{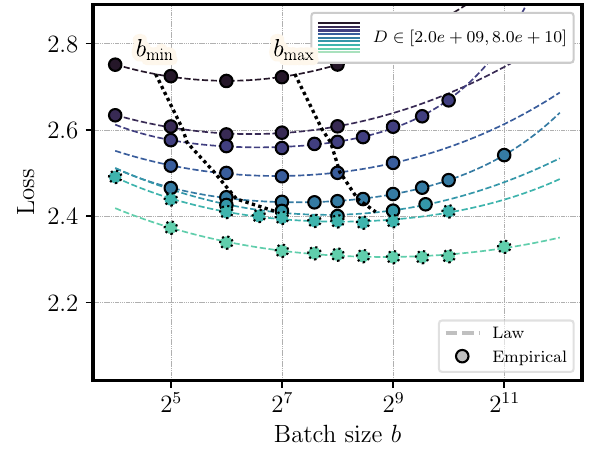}
    \end{subfigure}
    \caption{$N=268$M. \textbf{(Left)} Batch size range $[b_{\min}, b_{\max}]$ with $\epsilon$-suboptimal loss derived from law \eqref{eqn:suboptimal-bs-form} (with $\epsilon$ such that less than $5\%$ compute is wasted). Shaded area is obtained from fitting a power-law on the values of $b_{\min/\max}$ in-sample (solid lines). \textbf{(Right)} Empirical and predicted loss value across batch size $b$. Here, the predicted values are from the law \eqref{eqn:suboptimal-bs-form}, fitted separately for each $D$. Black dotted lines mark $b_{\min/\max}$ used for fitting the power-law $b_{\min/\max} \propto D^\nu$ on the left. Plots for other model sizes in \cref{sec:app:subopt-bs-li}.} 
    \label{fig:suboptimal-batch-size-two-stage}
\end{figure}

\begin{figure}[t]
    \centering
    \begin{subfigure}[b]{0.49\columnwidth}
        \includegraphics[width=0.99\linewidth]{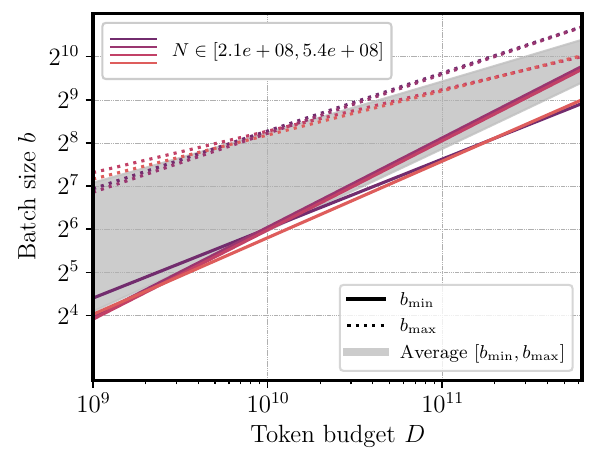}
        \caption{\li{} (sequence length $2048$)}
    \end{subfigure}
    \begin{subfigure}[b]{0.49\columnwidth}
        \includegraphics[width=0.99\linewidth]{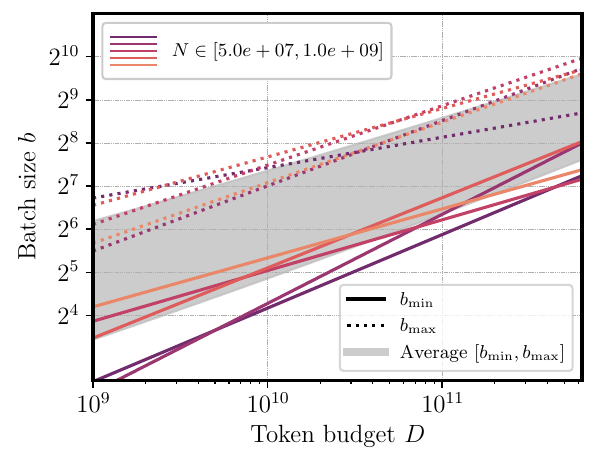}
        \caption{\oellm{} (sequence length $4096$)}
    \end{subfigure}
    \caption{Scaling of $\epsilon$-suboptimal batch size across model sizes, for \li{} dataset \textbf{(left)} and \oellm{} dataset \textbf{(right)}. The scaling of suboptimal batch sizes $[b_{\min}, b_{\max}]$ (grey area) is relatively consistent across the two datasets, after accounting for a factor of two due to the different sequence length.} 
    \label{fig:suboptimal-batch-size-all-ms}
\end{figure}

From \cref{fig:suboptimal-batch-size-two-stage}, we observe that this two-stage procedure -- as expected -- leads to a better fit across token budgets $D$. In particular, the fitted power-law on $b_{\min/\max}$ generalizes well beyond the token budgets used for fitting (\cref{fig:suboptimal-batch-size-two-stage}, left). See \cref{sec:app:subopt-bs-li,sec:app:subopt-bs-oellm} for additional model sizes, and for the \oellm{} dataset. When averaging the intervals of suboptimal batch sizes $[b_{\min}, b_{\max}]$ across model sizes (\cref{fig:suboptimal-batch-size-all-ms}), we observe a slightly narrowing trend for the \li{} dataset, and a relatively constant width for \oellm{}; except for this, the picture is overall similar across the two datasets, suggesting that the scaling behavior of suboptimal batch sizes with $D$ is relatively consistent.


\takeaway{The interval of $\epsilon$-suboptimal batch sizes can be modeled with a two-stage fitting procedure based from the three-term law; the scaling behavior generalizes well and is fairly consistent across model sizes and training setups. As rule of thumb, the interval of suboptimal batch sizes that corresponds to wasting at most 5\% of compute has roughly a width of $2^2$ (\cref{fig:suboptimal-batch-size-all-ms}).}


\subsection{Three-term Law and Critical Batch Size}\label{sec:cbs-analysis}

The notion of \emph{critical batch size} can be defined as follows: for a fixed target loss $\bar{\mathcal{L}}$, let 
$K_{\bar{\mathcal{L}}}(b)$ be the number of steps to reach loss $\bar{\mathcal{L}}$, as a function of the batch size $b$. As explained in the introduction, \citet{McCandlish2018} show that $K_{\bar{\mathcal{L}}}(b)$ decreases at much slower rate than inverse-linearly beyond a critical value of $b$, the so-called \emph{critical batch size}. This has an important practical consequence: training at the highest practically feasible batch size can be suboptimal if it exceeds the critical batch size. 

Empirically, \citet{Zhang2025} show that (i) critical batch size scales with compute under Chinchilla-optimal scaling of $(N,D)$, and (ii) this increase comes mostly from scaling up the token budget $D$. In particular, when $D$ is fixed, the function $K_{\bar{\mathcal{L}}}(b)$ is roughly the same across model sizes $N$.\footnote{Note that \citet{Zhang2025} operate in a slightly non-standard setup, as they use constant learning-rate schedule with weight averaging (vs.\ cosine schedule used in \citet{Li2025}).} Furthermore, \citet{Zhang2025} derive a theoretical model of critical batch size that captures the phenomena described above, however only for the very restricted setting of least-squares problems in the infinite-width limit.

Here, we show that the three-term law \eqref{eqn:three-term-law} can equally well describe the behaviour of $K_{\bar{\mathcal{L}}}(b)$ when scaling $D$ and/or $N$. For this, fix a target loss $\bar{\mathcal{L}}$ and denote $\tilde{E}(N):=E + \frac{A}{N^\alpha}$. From \eqref{eqn:three-term-law}, if $\bar{\mathcal{L}} > \tilde{E}(N) + \frac{CB}{M^\beta}$, the number of steps to reach $\bar{\mathcal{L}}$ is given by
\begin{align}\label{eqn:steps-to-loss}
    K_{\bar{\mathcal{L}}}(b) = \Big[ \frac{\bar{\mathcal{L}} - \tilde{E}(N)}{C} - \frac{B}{M^\beta}\Big] ^{-\frac{1}{\gamma}} =
    \Big[ \frac{\bar{\mathcal{L}} - \tilde{E}(N)}{C} - \frac{B}{Cs^\beta b^{\beta}}\Big] ^{-\frac{1}{\gamma}}.
\end{align}

We replicate the setting of \citet[Figure 1]{Zhang2025}, but plugging into \tttl{}:
\begin{enumerate}[label=(A\arabic*)]
    \item Scale up both $N$ and $D$ in the Chinchilla-optimal setting. 
    \item Fix $D=3.07$B and vary $N$ from $85$M to $1.2$B.
    \item Fix $N=302$M and vary $D$ in $[0.25,4]$ times the Chinchilla-optimal $D$.
\end{enumerate}

\begin{figure}[t]
    \centering
    \begin{subfigure}[b]{0.32\columnwidth}
        \includegraphics[width=0.99\linewidth]{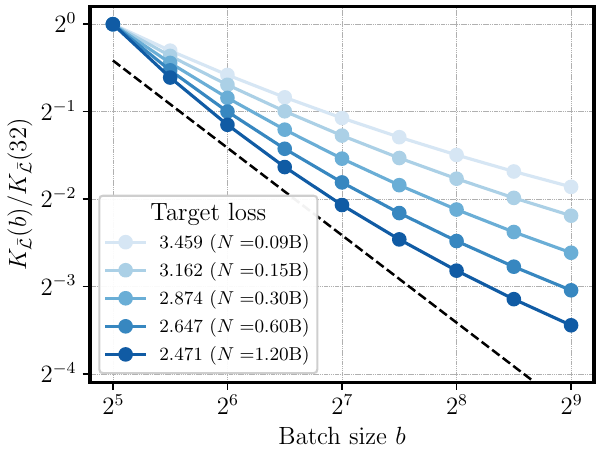}
        \caption{Chinchilla-optimal}
    \end{subfigure}
    \begin{subfigure}[b]{0.32\columnwidth}
        \includegraphics[width=0.99\linewidth]{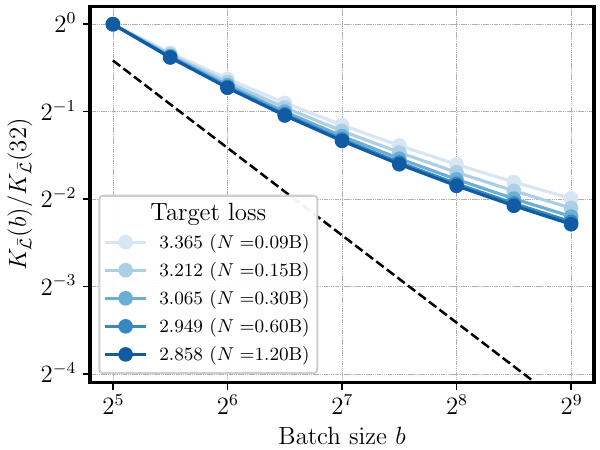}
        \caption{Fixed data $D=3.07$B}
    \end{subfigure}
    \begin{subfigure}[b]{0.32\columnwidth}
        \includegraphics[width=0.99\linewidth]{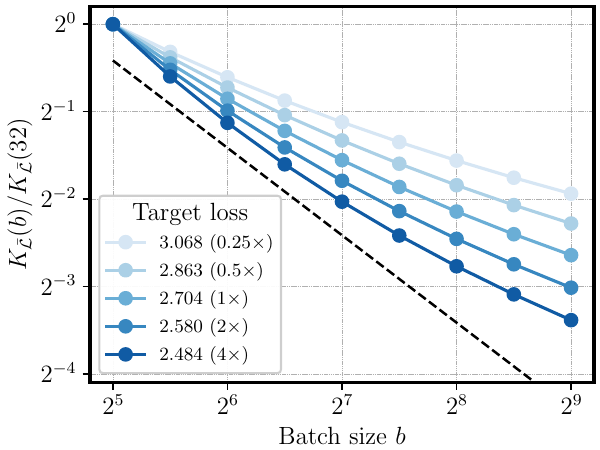}
        \caption{Fixed model $N=302$M}
    \end{subfigure}
    \caption{Under the three-term law, critical batch size changes with token budget $D$ \textbf{(right)}, but is almost invariant to changes of model size $N$ \textbf{(middle)}. This matches empirical findings, cf.\ \citet[Figure 1]{Zhang2025}.}
    \label{fig:cbs-analysis-li}
\end{figure}

For all three settings, given $(N,D)$ we compute the target loss $\bar{\mathcal{L}}$ from the law \eqref{epochai-law}. Then, we compute $K_{\bar{\mathcal{L}}}(b)$ according to \eqref{eqn:steps-to-loss}, using the parameters of \tttl{} previously fitted on the \li{} dataset. \cref{fig:cbs-analysis-li} confirms that the function $K_{\bar{\mathcal{L}}}(b)$ is almost invariant as we scale up model size $N$, but changes significantly if we scale up the token budget $D$.

\takeaway{Under the three-term law, the number of steps $K_{\bar{\mathcal{L}}}(b)$ to reach a target loss $\bar{\mathcal{L}}$, as a function of the batch size, is mostly invariant to scaling up $N$, but not to scaling up $D$. This matches the empirical results of \citet{Zhang2025}.}

As a consequence, the three-term law is a suitable model for $K_{\bar{\mathcal{L}}}(b)$ at large batch sizes, while \emph{at the same time} allowing for a non-trivial optimal batch size. This is in contrast to the theoretical models by \citet{McCandlish2018,Bergsma2025} and \citet{Zhang2025}, which can describe the critical batch size, but imply that the optimal batch size is one.

\paragraph{Comparison to other related models.} We would like to mention a different approach by \citet[Section 4.5]{Ruette2026}, which modifies the model of \citet{McCandlish2018} such that it also allows for optimal batch sizes larger than one. They propose the equation
\begin{align}\label{eqn:ruette-cbs-model}
    \Big(\Big[\frac{K}{K_{\min}}\Big]^{\alpha} - 1\Big)\Big(\Big[\frac{b}{b_{\min}}\Big]^{\alpha} - 1\Big) = 1.    
\end{align}
In comparison to the three-term law, the main conceptual difference is the setup of objective and constraints: \citet{Ruette2026} fix a target loss $\bar{\mathcal{L}}$, and minimize $D=bsK$ such that $\bar{\mathcal{L}}$ is reached, subject to \eqref{eqn:ruette-cbs-model}; for \tttl{}, we fix $D$, and minimize the final loss with respect to $b$ subject to $D=bsK$. Given that for a concrete training run it is much easier to fix $D$ than a target loss, the latter seems to be the more practicable approach.

\subsection{Back to the Chinchilla Form}\label{sec:back-to-chinchilla}
Having fitted the \tttl{}, we can compare its batch-size-optimal reduction from \eqref{eqn:three-term-optimal} to a Chinchilla-type law. To do so, we first fit the form \eqref{eqn:scaling-law} to the runs from \citet{Li2025} (only using optimal batch sizes); we use the same fitting procedure except that we set $\delta=10^{-5}$, which leads to a more stable fit. See \cref{sec:app:experiment-setup} in the Appendix for more details and ablations of this choice. 

Comparing this to the fitted parameters of the \tttl{} (see \cref{tab:back-to-chinchilla}), we can already see a rather big discrepancy; for example, with \tttl{} we obtain a much smaller value of $\tau$ (compared to $\beta$) as well as $E\approx0$.

Instead of simply comparing the parameter values, we can also compare the implied scaling behavior of both laws. In particular, the main goal of Chinchilla scaling laws is to determine how the optimal model size scales with compute $\mathcal{C}$. Assuming $\mathcal{C}=6ND$, from \eqref{eqn:scaling-law} we get $N^\star = \Big[\frac{\alpha A}{\beta B}\Big]^{\frac{1}{\alpha+\beta}} \big(\frac{\mathcal{C}}{6}\big)^\frac{\beta}{\alpha+\beta}$. For the three-term law, using \eqref{eqn:three-term-optimal}, we replace $(\beta,B)\to(\tau, \hat{B})$. \cref{fig:compute-optimal-model-size} shows that the implied compute-optimal scaling of $N^\star$ overlaps only for a relatively small interval of compute $\mathcal{C}$. 

\caveat{The three-term law \emph{can} be reduced to a Chinchilla-type law, however, its implied compute-optimal scaling is quite different to a direct fit of \eqref{eqn:scaling-law}. In particular, the implied scaling in $D$ is much smaller. This suggests that \tttl{} is not the most reliable instrument to describe compute-optimal allocation of $N$ and $D$.
}
This confirms the finding of \citet{Li2025a}, that the exact formulation of the scaling law can already impact the implied optimal model size. 

\begin{table}[ht]
\begin{minipage}[b]{0.45\linewidth}
\centering
\begin{tabular}{ | c | c | c | c |}
    \hline
    Parameter & Law \eqref{eqn:scaling-law} & \tttl{} \\ \hline \hline
    $E$ & 1.26 & 0.0 \\ \hline
    $A$ & 23.82 & 12.58 \\ \hline
    &&\\ [-1em]
    $B \mid \hat{B}$ & 4885.3 & 9.16 \\ \hline
    $\alpha$ & 0.166 & 0.132 \\ \hline
    $\beta \mid \tau$ & 0.336 & 0.079 \\ \hline
   \end{tabular}
   \vspace{3ex}
   \caption{Fitted parameter values when fitting a law of form \eqref{eqn:scaling-law} to \li{} dataset, and for \tttl{} after the reduction \eqref{eqn:three-term-optimal}.}
   \label{tab:back-to-chinchilla}
\end{minipage}\hfill
\begin{minipage}[b]{0.54\linewidth}
\centering
\includegraphics[width=0.8\linewidth]{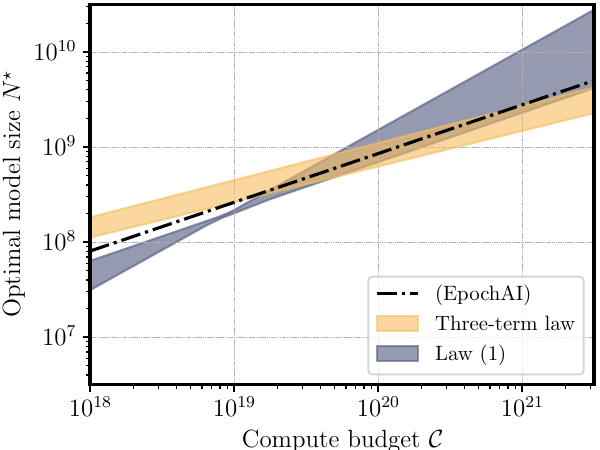}
\captionof{figure}{Compute-optimal model size.}
\label{fig:compute-optimal-model-size}
\end{minipage}
\end{table}

\section{Limitations}

We have already addressed some caveats in the discussions above. Here, we summarize the main limitations of the presented approach and how they could be resolved in future work:

\begin{itemize}
    \item While we have shown that the three-term law can be robustly fit for two different datasets (\li{} and \oellm{}), the quantitative results can be inconsistent to a degree which is minor (e.g.\ for optimal batch size scaling) or moderately high (e.g.\ impact of model size). It is not clear how well the reported scaling laws generalize to other training setups or tasks. Further, although more sophisticated scaling law formulations can in principle collapse back to the Chinchilla form, the resulting scaling can be quite different (\cref{sec:back-to-chinchilla}).
    \item While the three-term law explicitly models the batch size, we still need the optimal learning rate for each single combination of $(N,D,b)$; thus, despite our finding that the required amount of training runs can be reduced, the absolute number is still huge (before selecting the optimal learning rate, the \li{} dataset contains roughly 3000 runs). An interesting direction for future work would be to introduce the learning rate in the three-term law, possibly inspired again by the theoretical results from \citet{Shulgin2026} or similar works. However, given the previous limitation, it is unlikely that adding additional terms will alleviate the issue of consistency.    
    \item As we have seen, the three-term law alone is not predictive enough to infer the interval of $\epsilon$-suboptimal batch sizes. For the two-stage procedure we propose instead, we still require a relatively fine-grained batch size sweep.
    \item Optimal batch size scaling might be optimizer-dependent; in particular, it has been shown that the \texttt{Muon} optimizer \citep{Jordan2024} allows for larger batch sizes \citep{EssentialAI2025}. Investigating how switching the optimizer affects the fitted three-term law remains future work.
\end{itemize}

\section{Conclusion}

We have proposed a three-term scaling law that takes into account model size, training steps and batch size; the latter two explicitly model how the total amount of tokens is allocated. This formulation has natural advantages, bringing together Chinchilla-type and hyperparameter scaling laws, as well as tying it closely to theoretical results in stochastic optimization.

On a practical side, we have shown that our proposed law can be robustly fit even with incomplete batch size sweeps, thus largely reducing the number of training runs necessary to obtain scaling laws for the optimal batch size.

Second, our approach naturally allows to model suboptimal batch sizes, and we have derived their scaling with the total data budget. Finally, we have shown that the three-term law, in contrast to previous proposals, correctly describes the phenomenon of critical batch size, while at the same time allowing for non-trivial optimal batch size.

\section*{Acknowledgments}

Fabian Schaipp is supported by the French government under the
management of Agence Nationale de la Recherche as part of the “Investissements d’avenir” program,
reference ANR-19-P3IA-0001 (PRAIRIE 3IA Institute), and the
European Research Council Starting Grant DYNASTY – 101039676.

First, we would like to thank the authors of \citet{Li2025} for making all of their training runs public; without their dataset, this article would not have not been realized.

Second, many thanks go to Niccol\`{o} Ajroldi for compiling and providing access to the \oellm{} dataset, and to both Niccol\`{o} Ajroldi and Antonio Orvieto for their feedback and suggestions, which inspired some of the ideas of the paper.

Furthermore, this paper has benefited from discussions with Francis Bach, Alexander H\"{a}gele, Frederik Kunstner, Umut \c{S}im\c{s}ekli, and Adrien Taylor.

\bibliography{lib}

\clearpage

\appendix

\tableofcontents

\section{Experiments: Supplementary Material}\label{sec:app:experiments-supplementary}

\subsection{Details on Experimental Setup}\label{sec:app:experiment-setup}

\paragraph{Datasets.} Below is a short description and source for the two main datasets we use in the analysis. Within each dataset, the sequence length is the same across runs ($2048$ for \li{}, and $4096$ for \oellm{}). For the scaling laws, we select for each combination of $(N,b,D)$ (model size, batch size, token budget) the learning rate that obtains smallest final loss.

\noindent\rule{\textwidth}{1pt}
\begin{description}[listparindent=0pt,leftmargin=1.0em,]
	\item[\li] \hfill \citep{Li2025}~\\[.5em]
	Training logs for different model sizes, token budgets, batch sizes, and learning rates. We use their smoothened loss and filter on dense models (no MoEs).
	
	
    \begin{tabular}{@{}ll@{}}
		Source & \small \url{https://wandb.ai/billzid/predictable-scale}
	\end{tabular}

	\item[\oellm] \hfill \citep{OpenEuroLLM2026}~\\[.5em]
    This is an unpublished dataset of training runs executed by the \href{www.openeurollm.eu}{OpenEuroLLM initiative}. Additional details on the training setup will be provided upon its release 
	by the OpenEuroLLM initiative.

\end{description}
\noindent\rule{\textwidth}{1pt}

\begin{table}[H]
    \centering
    \caption{Overview of the datasets used for fitting our scaling laws. Here, we report the ranges of $(N,M,K)$ on the union of train and validation set. See also \cref{fig:data-overview-li}.}
    \begin{tabular}{|c|cc|}
        \hline
        \textbf{Name} & \li{} &  \oellm{} \\[1pt]
        \hline \hline
        Samples (train/val) & (246/54) & (191/21) \\
        Range of $N$ & $[59, 1073] ~(\cdot 10^6)$ & $[50, 1000]~(\cdot 10^6)$ \\
        Range of $M$ & $[0.032,4.2] ~(\cdot 10^6)$ & $[0.65,4.19] ~(\cdot 10^6)$ \\
        Range of $K$ & $[0.9, 1525] ~(\cdot 10^3)$ & $[7.1,1144.4] ~(\cdot 10^3)$ \\
        Range of $\mathcal{L}$ & $[2.12, 3.14]$ & $[2.08,2.93]$ \\
        \hline
    \end{tabular}
    \label{tab:dataset-overview}
\end{table}

\paragraph{Fitting methodology.} We first describe the procedure for fitting the scaling laws. The training set is split into five parts of equal size. For each law, we then fit the same law on each leave-one-out split of datapoints (five-fold cross-validation). This allows for more robustness to outliers and to quantify uncertainty for each fitted parameter.

For the fitting procedure, we use the same Huber loss function as \citet{Besiroglu2024}: for true loss values $\mathcal{L}_{\text{true}}$ and predicted loss values $\hat{\mathcal{L}}$, we minimize
\begin{align*}
    \sum_{i} \mathcal{H}_\delta\big(\log(\mathcal{L}^{(i)}_{\text{true}}) - \log(\hat{\mathcal{L}}^{(i)}\big), \quad 
    \mathcal{H}_\delta(z) = \begin{cases}
        \frac12 \delta^2, \quad &|z| \leq \delta, \\
        \delta(|z|-\frac12 \delta), \quad &\text{else}.
    \end{cases}
\end{align*}
We use the \texttt{minimize} method from \texttt{scipy.optimize} together with the \texttt{L-BFGS-B} optimizer.

\paragraph{Choice of $\delta$.} For the Huber loss, we set $\delta=10^{-3}$ for the \ttl{} and \tttl{} scaling laws, which is the standard value used also by \citet{Besiroglu2024}. We run an ablation on this choice, setting  $\delta =10^{-5}$ instead, see \cref{sec:app:ablation-delta} for details.

\paragraph{Initialization.} For each single fit, we minimize the Huber loss at $n_{\text{init}}$ different initializations  with \texttt{L-BFGS}, and select the solution that results in the smallest objective function. By default, we use a grid of ten values for each of $(\alpha, \beta, \gamma, E)$ and two values for each of $(A,B,C)$. That is, for a \ttl{} (see \eqref{eqn:two-term-law}) we have $n_{\text{init}} =  10^3 \cdot 2^2=4000$; for a \tttl{} (see \eqref{eqn:three-term-law}) this becomes computationally intensive, and we therefore randomly select $5000$ starting points from the grid. 

\paragraph{Evaluation.} After fitting, we predict loss values by averaging over the predictions of each of the five individually fit models (\emph{cross-validation ensemble}). In \cref{tab:bs-step-laws-li,tab:bs-step-laws-oellm}, we report the in-sample mean-absolute deviation (MAD) of the predicted and true loss values.

\subsection{Additional Observations}\label{sec:app:additional-observations}
This section lists changes of the fitting technique that we have tried (usually only on the \li{} dataset), which however do not lead to better or significantly different results.
\begin{enumerate}[label=(\Roman*)]
    \item Due to the almost-zero values of the parameter $E$ in our laws, we try to enforce larger values by adding the regularization term $\frac{\lambda}{2}(\log{E})^2$. With $\lambda=10^{-3}$, this leads to different coefficients, in particular $E=0.729$; in terms of the other analyses, for example optimal batch size scaling, this has no major impact.
    \item Providing the gradient of \eqref{eqn:three-term-law} to the \texttt{L-BFGS} solver has no major impact on results.
    \item Using \texttt{BFGS} instead of \texttt{L-BFGS} has no major impact on results (while being significantly slower for fitting).
\end{enumerate}


\subsection{Scaling Law Coefficients}\label{sec:app:coefficients}

\begin{table}[H]
\centering
\caption{\li{} dataset: Scaling law coefficients for \ttl{} (see \eqref{eqn:two-term-law}) across model sizes, as well as for \tttl{} (see \eqref{eqn:three-term-law}). \ttl{} has no parameters $(A, \alpha)$, see \eqref{eqn:two-term-law}. We report average value across subsampled fits, with the standard deviation in brackets. MAD refers to the in-sample mean absolute deviation of the predicted (with cross-validation ensemble) to the true loss values.}
\label{tab:bs-step-laws-li}
\begin{tabular}{|c|cccc|ccc||cc|}
\hline
Model size $N$ & $E$ & $A$ & $B$ & $C$ & $\alpha$ & $\beta$ & $\gamma$ & Samples & MAD \\
\hline \hline
60M & $\underset{(0.31)}{0.705}$ & - & $\underset{(0.15)}{4.56}$ & $\underset{(1.4)}{7.26}$ & - & $\underset{(0.018)}{0.0864}$ & $\underset{(0.029)}{0.278}$ & 13 & 0.00802 \\[10pt]
120M & $\underset{(0.039)}{0.0395}$ & - & $\underset{(0.22)}{5.24}$ & $\underset{(1.7)}{8.22}$ & - & $\underset{(0.0082)}{0.0752}$ & $\underset{(0.044)}{0.274}$ & 13 & 0.0089 \\[10pt]
215M & $\underset{(0.38)}{0.481}$ & - & $\underset{(0.44)}{4.66}$ & $\underset{(0.6)}{5.01}$ & - & $\underset{(0.033)}{0.0983}$ & $\underset{(0.028)}{0.204}$ & 40 & 0.0123 \\[10pt]
268M & $\underset{(0.17)}{0.959}$ & - & $\underset{(0.3)}{4.04}$ & $\underset{(0.23)}{4.33}$ & - & $\underset{(0.019)}{0.113}$ & $\underset{(0.011)}{0.201}$ & 60 & 0.0143 \\[10pt]
429M & $\underset{(0.22)}{0.763}$ & - & $\underset{(0.73)}{5.21}$ & $\underset{(0.25)}{3.47}$ & - & $\underset{(0.027)}{0.151}$ & $\underset{(0.018)}{0.133}$ & 55 & 0.0115 \\[10pt]
537M & $\underset{(0.1)}{0.198}$ & - & $\underset{(0.55)}{5.05}$ & $\underset{(0.1)}{3.67}$ & - & $\underset{(0.02)}{0.13}$ & $\underset{(0.015)}{0.108}$ & 40 & 0.00962 \\[10pt]
1074M & $\underset{(2.7e-07)}{1.37e-07}$ & - & $\underset{(0.35)}{5.63}$ & $\underset{(0.19)}{4.3}$ & - & $\underset{(0.0087)}{0.126}$ & $\underset{(0.008)}{0.125}$ & 25 & 0.0111 \\[10pt]
\hline \hline
Three-term & $\underset{(2.2e-11)}{1.08e-11}$ & $\underset{(0.59)}{12.6}$ & $\underset{(0.2)}{4.9}$ & $\underset{(0.15)}{4.27}$ & $\underset{(0.0043)}{0.132}$ & $\underset{(0.008)}{0.139}$ & $\underset{(0.0074)}{0.182}$ & 246 & 0.0159 \\[10pt]
\hline
\end{tabular}
\end{table}
\begin{table}[H]
\centering
\caption{Same as \cref{tab:bs-step-laws-li}, but for \oellm{} dataset.}
\label{tab:bs-step-laws-oellm}
\begin{tabular}{|c|cccc|ccc||cc|}
\hline
Model size $N$ & $E$ & $A$ & $B$ & $C$ & $\alpha$ & $\beta$ & $\gamma$ & Samples & MAD \\
\hline \hline
50M & $\underset{(0.048)}{2.3}$ & - & $\underset{(0.42)}{2.26}$ & $\underset{(0.38)}{2.27}$ & - & $\underset{(0.026)}{0.161}$ & $\underset{(0.026)}{0.2}$ & 43 & 0.00654 \\[10pt]
130M & $\underset{(0.11)}{1.79}$ & - & $\underset{(0.76)}{3.45}$ & $\underset{(0.7)}{2.92}$ & - & $\underset{(0.028)}{0.164}$ & $\underset{(0.04)}{0.184}$ & 39 & 0.00798 \\[10pt]
300M & $\underset{(0.13)}{1.16}$ & - & $\underset{(0.21)}{3.07}$ & $\underset{(0.056)}{2.51}$ & - & $\underset{(0.018)}{0.113}$ & $\underset{(0.007)}{0.138}$ & 38 & 0.00656 \\[10pt]
600M & $\underset{(0.021)}{1.09}$ & - & $\underset{(0.36)}{4.41}$ & $\underset{(0.057)}{2.58}$ & - & $\underset{(0.011)}{0.159}$ & $\underset{(0.0085)}{0.122}$ & 36 & 0.00562 \\[10pt]
1000M & $\underset{(0.12)}{1.02}$ & - & $\underset{(2)}{5.17}$ & $\underset{(0.16)}{2.8}$ & - & $\underset{(0.042)}{0.169}$ & $\underset{(0.012)}{0.128}$ & 35 & 0.00718 \\[10pt]
\hline \hline
Three-term & $\underset{(0.23)}{0.264}$ & $\underset{(24)}{180}$ & $\underset{(0.059)}{2.62}$ & $\underset{(0.16)}{2.73}$ & $\underset{(0.0077)}{0.292}$ & $\underset{(0.017)}{0.0705}$ & $\underset{(0.011)}{0.156}$ & 191 & 0.0128 \\[10pt]
\hline
\end{tabular}
\end{table}
\subsection{Fitting \ttl{} and \tttl{} on \oellm{} Dataset}\label{sec:app:oellm-results}
\begin{figure}[H]
    \centering
    \begin{subfigure}[b]{0.49\columnwidth}
        \includegraphics[width=0.99\linewidth]{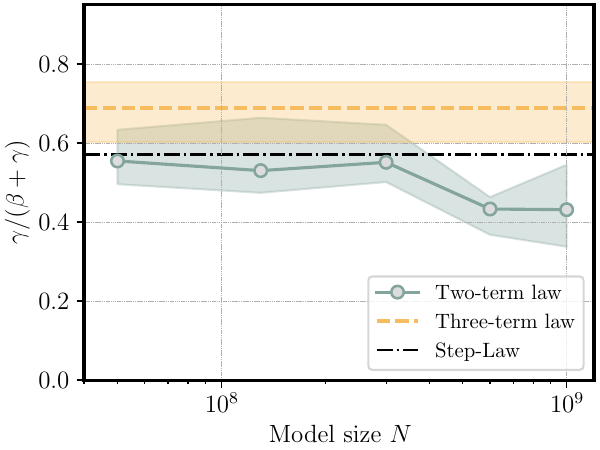}
    \end{subfigure}
    \begin{subfigure}[b]{0.49\columnwidth}
        \includegraphics[width=0.99\linewidth]{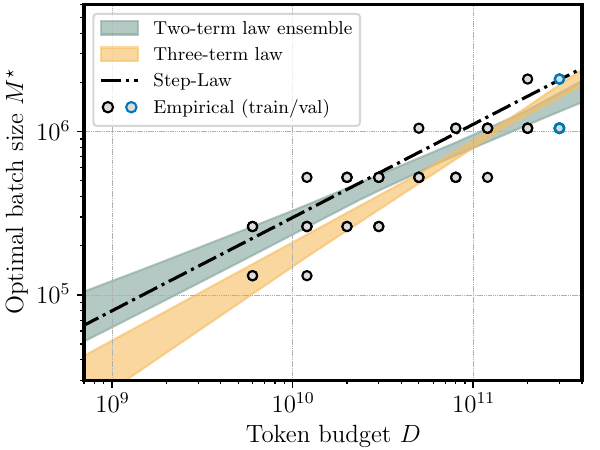}
    \end{subfigure}
    \caption{Same as \cref{fig:scaling-law-comparison-li}, but for \oellm{} dataset.}
    \label{fig:scaling-law-comparison-oellm}
\end{figure}
\begin{figure}[H]
    \centering
    \begin{subfigure}[b]{0.49\columnwidth}
        \includegraphics[width=0.99\linewidth]{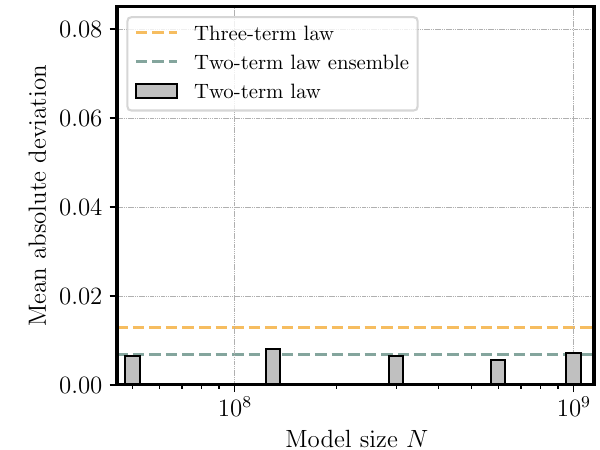}
    \end{subfigure}
    \begin{subfigure}[b]{0.49\columnwidth}
        \includegraphics[width=0.99\linewidth]{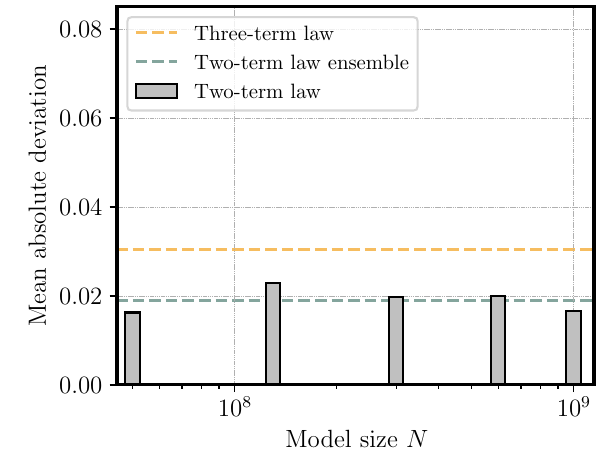}
    \end{subfigure}
    \caption{Same as \cref{fig:mad-comparison-li}, but for \oellm{} dataset. \textbf{(Left)} Training set, \textbf{(right)} validation set.}
    \label{fig:mad-comparison-oellm}
\end{figure}
%

%
\subsection{Ablation on Value of $\delta$}\label{sec:app:ablation-delta}
When fitting the standard Chinchilla form \eqref{eqn:scaling-law} to the \li{} dataset with $\delta = 10^{-3}$, we observe that the variance in the parameter estimates appears to be unbalanced; this is likely due to the rather small range of $N$ in the dataset, which has been reported to cause issues for fitting scaling laws \citep{Li2025a}. We find that using $\delta = 10^{-5}$ fixes this. Hence, we perform an ablation with $\delta=10^{-5}$ for the two-term and three-term scaling laws, to check whether the value of $\delta$ impacts also the fit of those laws.

Below we show the results of \cref{sec:experiments:compare-2tl-3tl}, but using $\delta=10^{-5}$ instead of $\delta=10^{-3}$. In short, for \tttl{} we observe almost identical results, albeit with slightly higher variance for the scaling of $M^\star$. For \ttl{}, we observe a slightly worse MAD when using $\delta=10^{-5}$, as well as higher variance for the coefficients $(\gamma, \beta)$ as well as the scaling of $M^\star$.

Overall, the choice of $\delta$ does not have big impact on the conclusion of \cref{sec:experiments:compare-2tl-3tl}, with $\delta=10^{-3}$ being slightly preferable.
\begin{figure}[H]
    \centering
    \begin{subfigure}[b]{0.49\columnwidth}
        \includegraphics[width=0.99\linewidth]{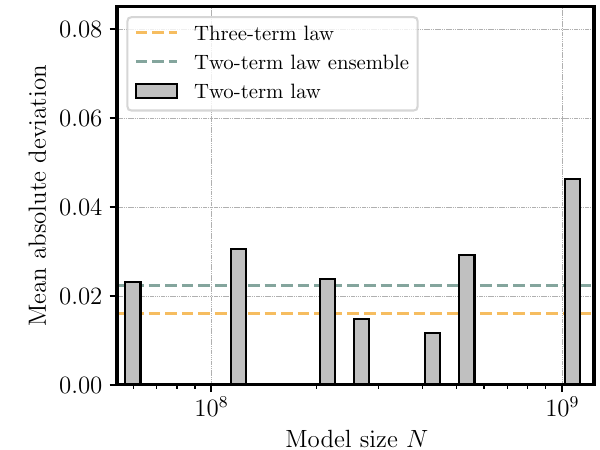}
    \end{subfigure}
    \begin{subfigure}[b]{0.49\columnwidth}
        \includegraphics[width=0.99\linewidth]{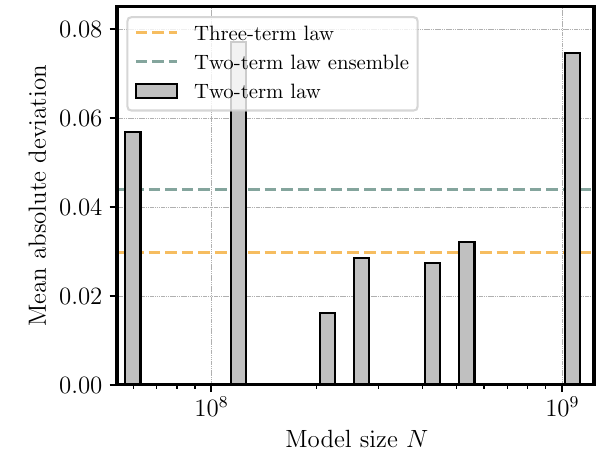}
    \end{subfigure}
    \caption{Same as \cref{fig:mad-comparison-li}, but with $\delta=10^{-5}$.}
    \label{fig:mad-comparison-li-ablate-delta}
\end{figure}
\begin{figure}[H]
    \centering
    \begin{subfigure}[b]{0.49\columnwidth}
        \includegraphics[width=0.99\linewidth]{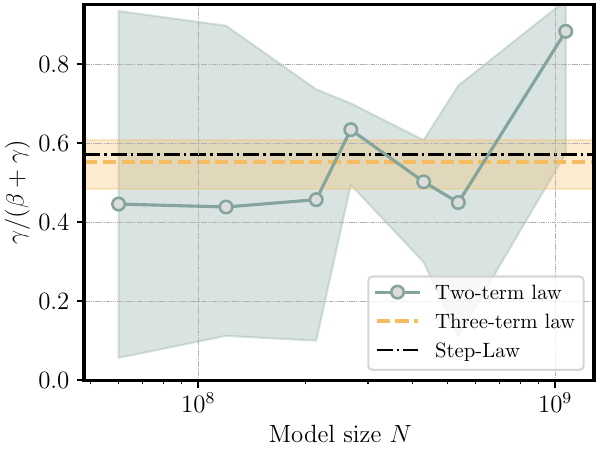}
    \end{subfigure}
    \begin{subfigure}[b]{0.49\columnwidth}
        \includegraphics[width=0.99\linewidth]{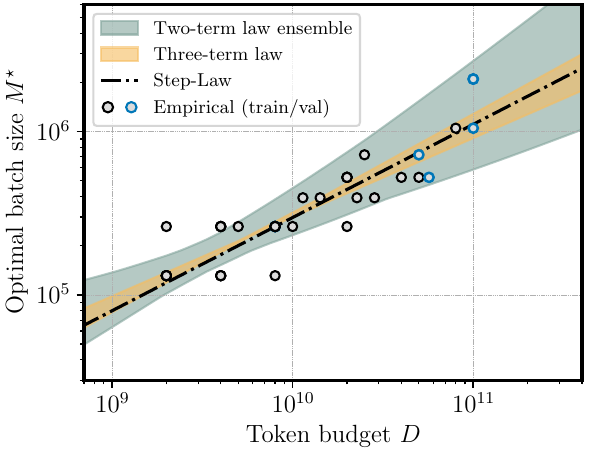}
    \end{subfigure}
    \caption{Same as \cref{fig:scaling-law-comparison-li}, but with $\delta=10^{-5}$.}
    \label{fig:scaling-law-comparison-li-ablate-delta}
\end{figure}
\clearpage
\subsection{Additional Plots}\label{sec:app:plots}
\subsubsection{\li{} Dataset Overview}\label{sec:app:plots-li}
\begin{figure}[H]
    \centering
    \includegraphics[width=0.99\linewidth]{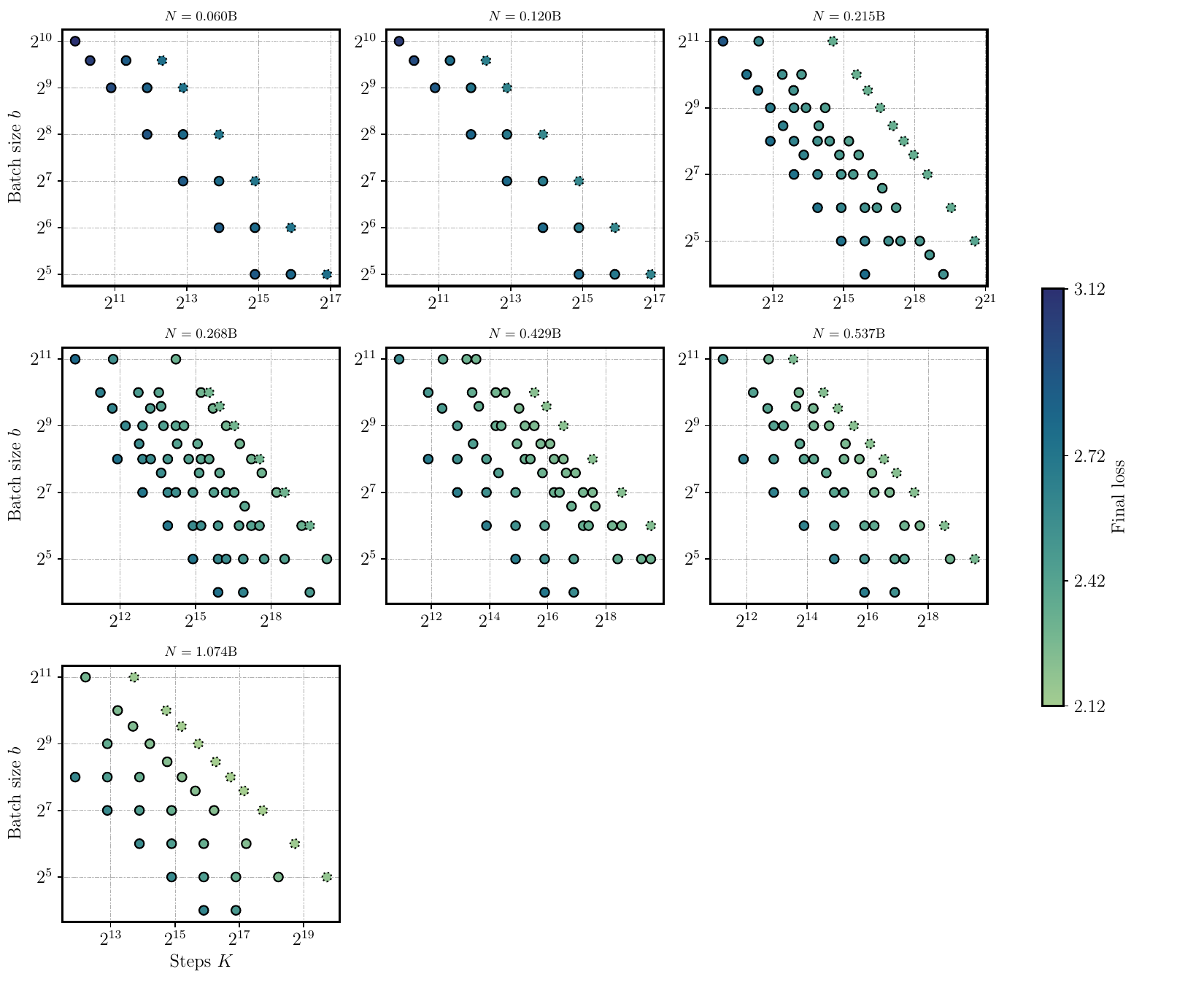}
    \caption{Overview of the \li{} dataset used for fitting scaling laws. Dots with dashed border are part of the validation set.}
    \label{fig:data-overview-li}
\end{figure}
\clearpage
\begin{figure}[H]
    \centering
    \includegraphics[width=0.99\linewidth]{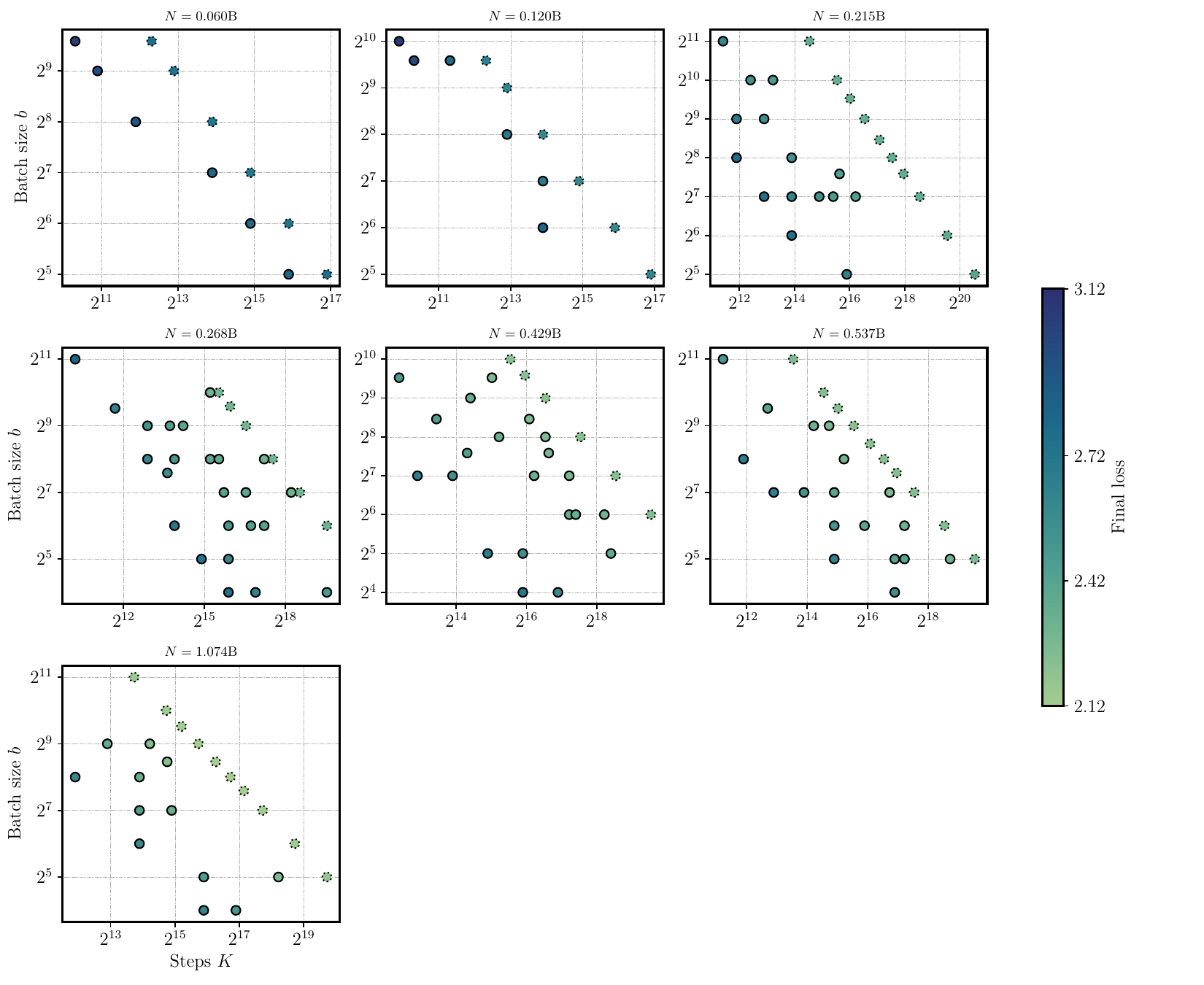}
    \caption{Illustration of the reduced dataset used in \cref{sec:saving-compute}, with 3 batch size values $b$ per sweep. See also caption of \cref{fig:data-overview-li}; here, the difference is that for each value of $(N,D)$ we use only three different batch sizes (lying on a diagonal) for fitting, while the validation set remains the same as before.}
    \label{fig:data-overview-li-mask3}
\end{figure}
\clearpage
\begin{figure}[H]
    \centering
    \includegraphics[width=0.99\linewidth]{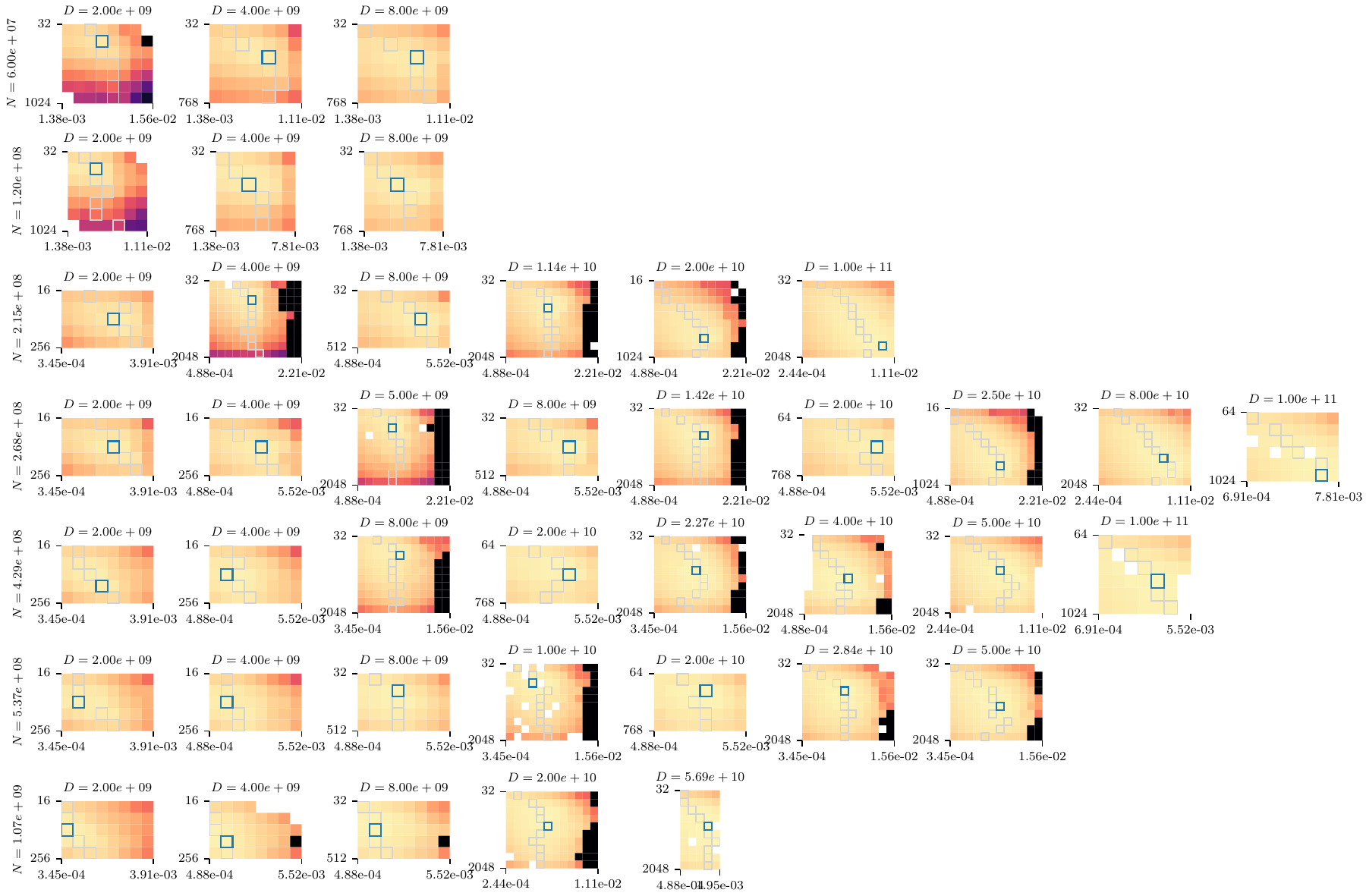}
    \caption{Overview of the full \li{} dataset (before learning-rate selection). Each heatmap represents the final loss over a grid of batch size $b$ ($y$-axis) and learning rate $\eta$ ($x$-axis) for a single combination of $(N,D)$. Blue squares mark the optimal combination of $(\eta, b)$, gray squares mark optimal learning rate for the given row of batch size. Note that most marked squares do not lie on the border, therefore indicating that the sweep is sufficiently extensive.}
    \label{fig:heatmaps-li}
\end{figure}


\subsubsection{Suboptimal Batch Size Scaling: \li{}}\label{sec:app:subopt-bs-li}
\begin{figure}[H]
    \centering
    \begin{subfigure}[b]{0.49\columnwidth}
        \includegraphics[width=0.99\linewidth]{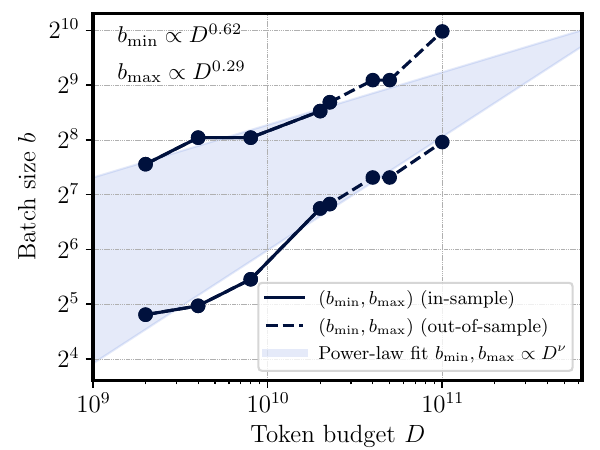}
    \end{subfigure}
    \begin{subfigure}[b]{0.49\columnwidth}
        \includegraphics[width=0.99\linewidth]{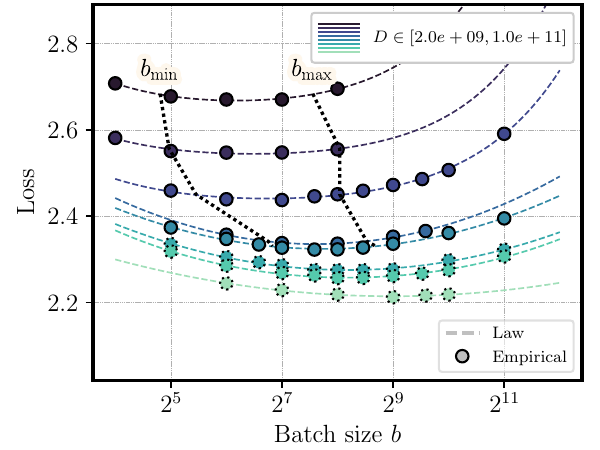}
    \end{subfigure}
    \caption{Same as \cref{fig:suboptimal-batch-size-two-stage}, but for $N=429$M.} 
    \label{fig:suboptimal-batch-size-two-stage-2}
\end{figure}
\begin{figure}[H]
    \centering
    \begin{subfigure}[b]{0.49\columnwidth}
        \includegraphics[width=0.99\linewidth]{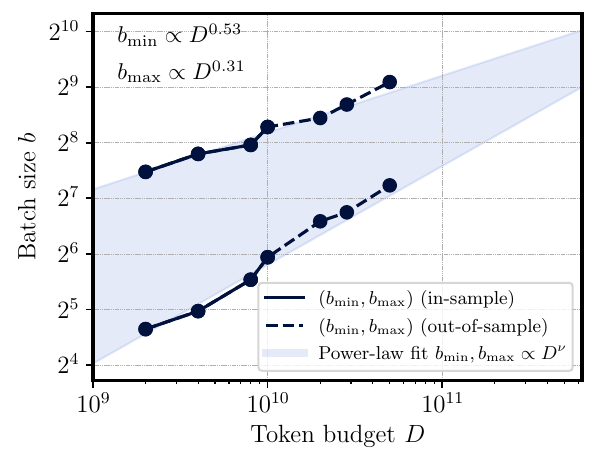}
    \end{subfigure}
    \begin{subfigure}[b]{0.49\columnwidth}
        \includegraphics[width=0.99\linewidth]{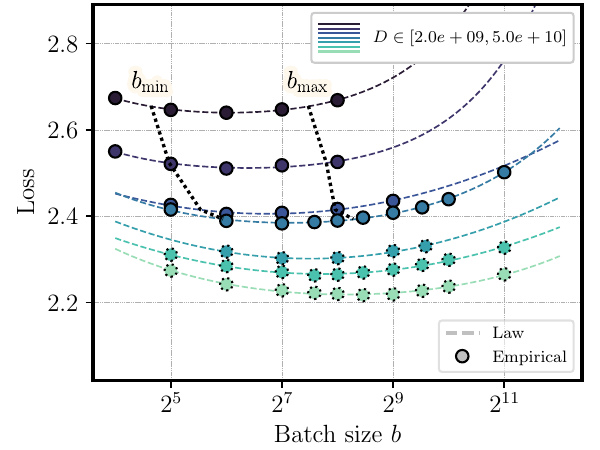}
    \end{subfigure}
    \caption{Same as \cref{fig:suboptimal-batch-size-two-stage}, but for $N=537$M.} 
    \label{fig:suboptimal-batch-size-two-stage-3}
\end{figure}
\subsubsection{Suboptimal Batch Size Scaling: \oellm{}}\label{sec:app:subopt-bs-oellm}
\begin{figure}[H]
    \centering
    \begin{subfigure}[b]{0.49\columnwidth}
        \includegraphics[width=0.99\linewidth]{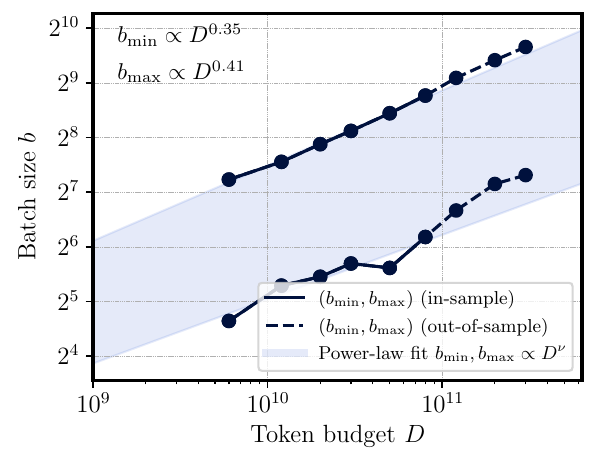}
    \end{subfigure}
    \begin{subfigure}[b]{0.49\columnwidth}
        \includegraphics[width=0.99\linewidth]{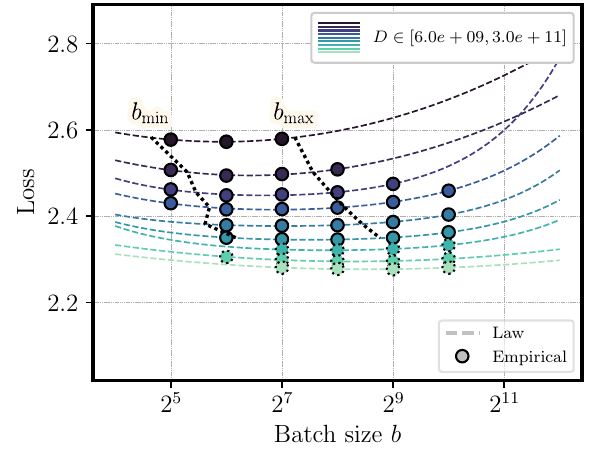}
    \end{subfigure}
    \caption{Same as \cref{fig:suboptimal-batch-size-two-stage}, but for \oellm{} dataset with $N=300$M.} 
    \label{fig:suboptimal-batch-size-two-stage-oellm}
\end{figure}
\begin{figure}[H]
    \centering
    \begin{subfigure}[b]{0.49\columnwidth}
        \includegraphics[width=0.99\linewidth]{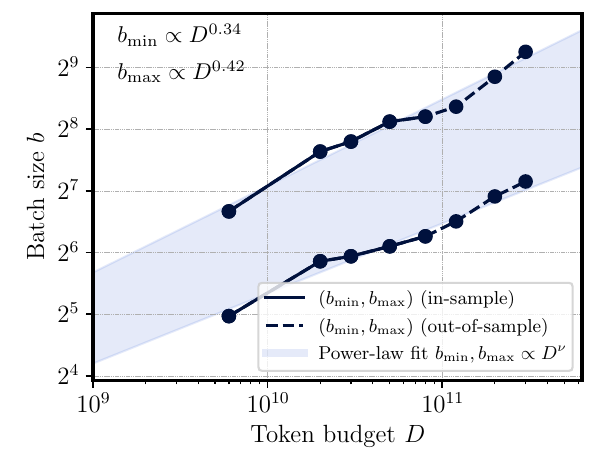}
    \end{subfigure}
    \begin{subfigure}[b]{0.49\columnwidth}
        \includegraphics[width=0.99\linewidth]{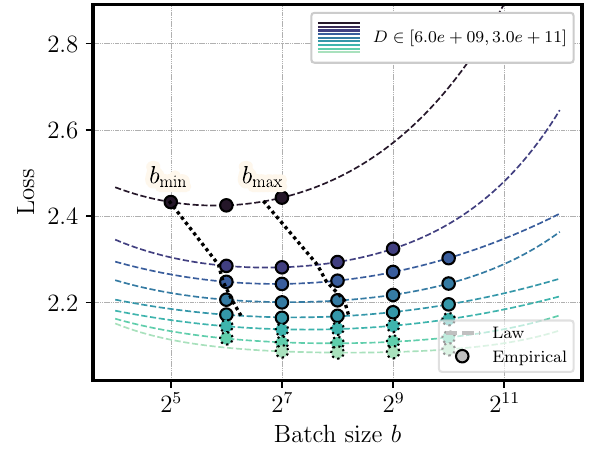}
    \end{subfigure}
    \caption{Same as \cref{fig:suboptimal-batch-size-two-stage}, but for \oellm{} dataset with $N=1$B.} 
    \label{fig:suboptimal-batch-size-two-stage-oellm-2}
\end{figure}
\subsubsection{Reduced Dataset Fit for \oellm{}}\label{sec:app:reduced-fit-oellm}
\begin{figure}[H]
    \centering
    \begin{subfigure}[b]{0.49\columnwidth}
        \includegraphics[width=0.99\linewidth]{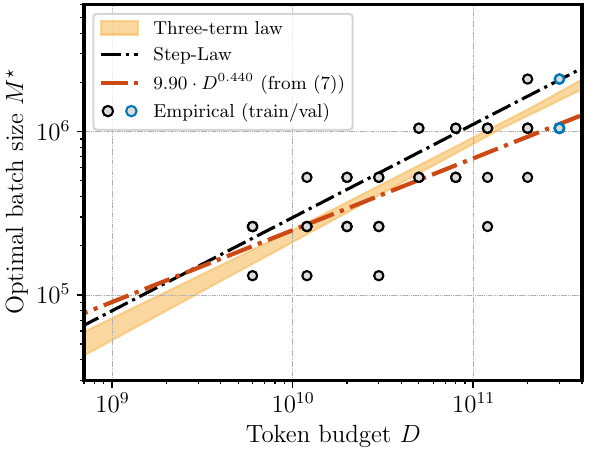}
        \caption{Dataset reduced to 63\%}
    \end{subfigure}
    \begin{subfigure}[b]{0.49\columnwidth}
        \includegraphics[width=0.99\linewidth]{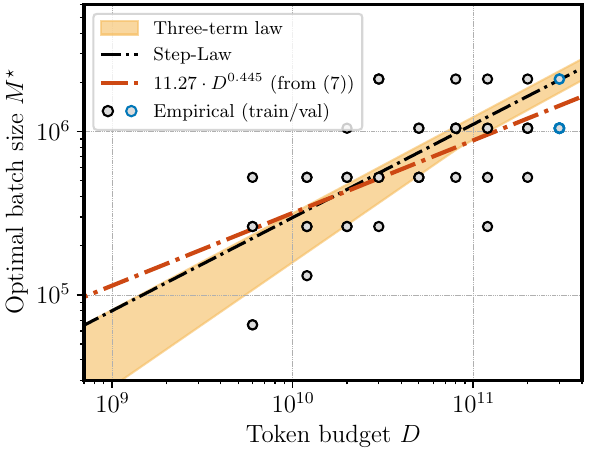}
        \caption{Dataset reduced to 42\%}
    \end{subfigure}
    \caption{Same as \cref{fig:masked-optimal-bs-scaling}, but for \oellm{} dataset. Fitting on a reduced dataset, with only 3 values of $b$ per sweep \textbf{(left)} and 2 values \textbf{(right)}.
    } 
    \label{fig:masked-optimal-bs-scaling-oellm}
\end{figure}

\end{document}